\documentclass{article} 
\usepackage{nips12submit_e,times}

\usepackage{tikz}

\usepackage{ upgreek }
\usepackage{verbatim} 

\usepackage{amssymb}
\usepackage{graphicx}
\usepackage{url}
\usepackage{verbatim} 
\usepackage{color}
\usepackage{wrapfig}

\usepackage[numbers]{natbib}
\usepackage{array}
\usepackage{relsize}
\usepackage{multirow}
\usepackage{amsmath}

\usepackage[font=small,format=plain,labelfont=bf,up]{caption}

 \belowdisplayskip \abovedisplayskip

\newlength{\sectionReduceTop}
\newlength{\sectionReduceBot}
\newlength{\subsectionReduceTop}
\newlength{\subsectionReduceBot}
\newlength{\abstractReduceTop}
\newlength{\abstractReduceBot}
\newlength{\captionReduceTop}
\newlength{\captionReduceBot}
\newlength{\subsubsectionReduceTop}
\newlength{\subsubsectionReduceBot}

\newlength{\horSkip}
\newlength{\verSkip}

\newlength{\figureHeight}
\newlength\savedwidth
\newcommand\whline[1]{\noalign{\global\savedwidth\arrayrulewidth
                               \global\arrayrulewidth #1} %
                      \hline
                      \noalign{\global\arrayrulewidth\savedwidth}}

\DeclareMathOperator*{\argmax}{argmax}

\newcommand{\M}{{M}}             
\newcommand{\x}{{\mathbf x}}     
\newcommand{\xh}[1]{{\mathbf x^{h_{#1}}}}     
\newcommand{\xs}[1]{{x_{#1}}}    
\newcommand{\y}{{\mathbf y}}     
\newcommand{\yh}[1]{{\mathbf y^{h_{#1}}}}     
\newcommand{\yhhat}[1]{{\mathbf {\hat{y}}^{h_{#1}}}}     
\newcommand{\ys}[1]{{\mathbf y_{#1}}}    
\newcommand{\ysc}[2]{{y_{#1}^{#2}}}    

\newcommand{\zsc}[2]{{z_{#1}^{#2}}}    

\newcommand{\fo}[1]{{\phi_o({#1})}}      
\newcommand{\fs}[1]{{\phi_a({#1})}}      
\newcommand{\fe}[3]{{\phi_{#1}(#2,#3)}}
\newcommand{\w}{{\mathbf w}}           
\newcommand{\wh}[1]{{\mathbf w^{h_{#1}}}}           
\newcommand{\wo}[1]{{w_o^{#1}}}        
\newcommand{\ws}[1]{{w_a^{#1}}}        
\newcommand{\we}[3]{{w_{#1}^{#2#3}}}   

\newcommand{\df}[3]{{f_{#3}(#1,#2)}}   
\newcommand{\dg}[3]{{g_{#3}(#1,#2)}}   
\newcommand{\loss}[2]{{\Delta(#1,#2)}}   

\renewcommand{\Re}{\mathbb{R}}

 \newcommand{\todo}[1]{\textcolor{blue}{\textbf{#1}}}

\title{Human Activity Learning using Object Affordances from RGB-D Videos}

\author{
Hema Swetha Koppula, Rudhir Gupta and Ashutosh Saxena\\
Department of Computer Science\\
Cornell University\\
Ithaca, NY 14850 \\
\texttt{\{hema,rg495,asaxena\}@cs.cornell.edu} \\
}

\nipsfinalcopy 

\begin{document}

\maketitle

\vskip -.1in

\begin{abstract} 
\vspace*{\abstractReduceBot}
Human activities comprise several sub-activities performed
in a sequence and involve interactions with various objects.  This makes
reasoning about the object affordances a central task for activity recognition.
In this work, we consider the problem of jointly labeling the object affordances and human 
activities from {RGB-D} videos. 
We frame the problem as a 
Markov Random Field where the nodes represent  objects and 
 sub-activities, and the edges represent the relationships between object 
affordances, their relations with sub-activities, and their evolution over
time. We formulate the learning problem using a structural
SVM approach, where labeling over various alternate temporal 
segmentations are considered as latent variables. We tested our method
on a dataset comprising 120 activity videos collected from four subjects,
and obtained an end-to-end precision of 
81.8\% and  recall of 80.0\% for labeling the activities.
\end{abstract}




\vspace{\sectionReduceTop}
 \section{Introduction}
 \vspace*{\sectionReduceBot}
 In this paper, we present a learning algorithm that takes as input an RGB-D video 
 (obtained from an inexpensive sensor such as Microsoft Kinect), and identifies
 the human activities taking place over long time periods (e.g., see Fig.~\ref{fig:modelfigure}).  
Most prior work in human activity detection has focussed on activity detection from still images
or from 2D videos. Estimating the human pose is the primary focus of these works, and they consider
 activities taking place over shorter time scales (see Section~\ref{sec:relatedwork}).
Having access to a 3D camera, which provides RGB-D videos, enables us to robustly 
 estimate human poses and use this information for  learning complex human activities. 


 


\begin{figure}[t!]
\vskip -.1in
\centering
\includegraphics[width=.16\linewidth,height=0.6in]{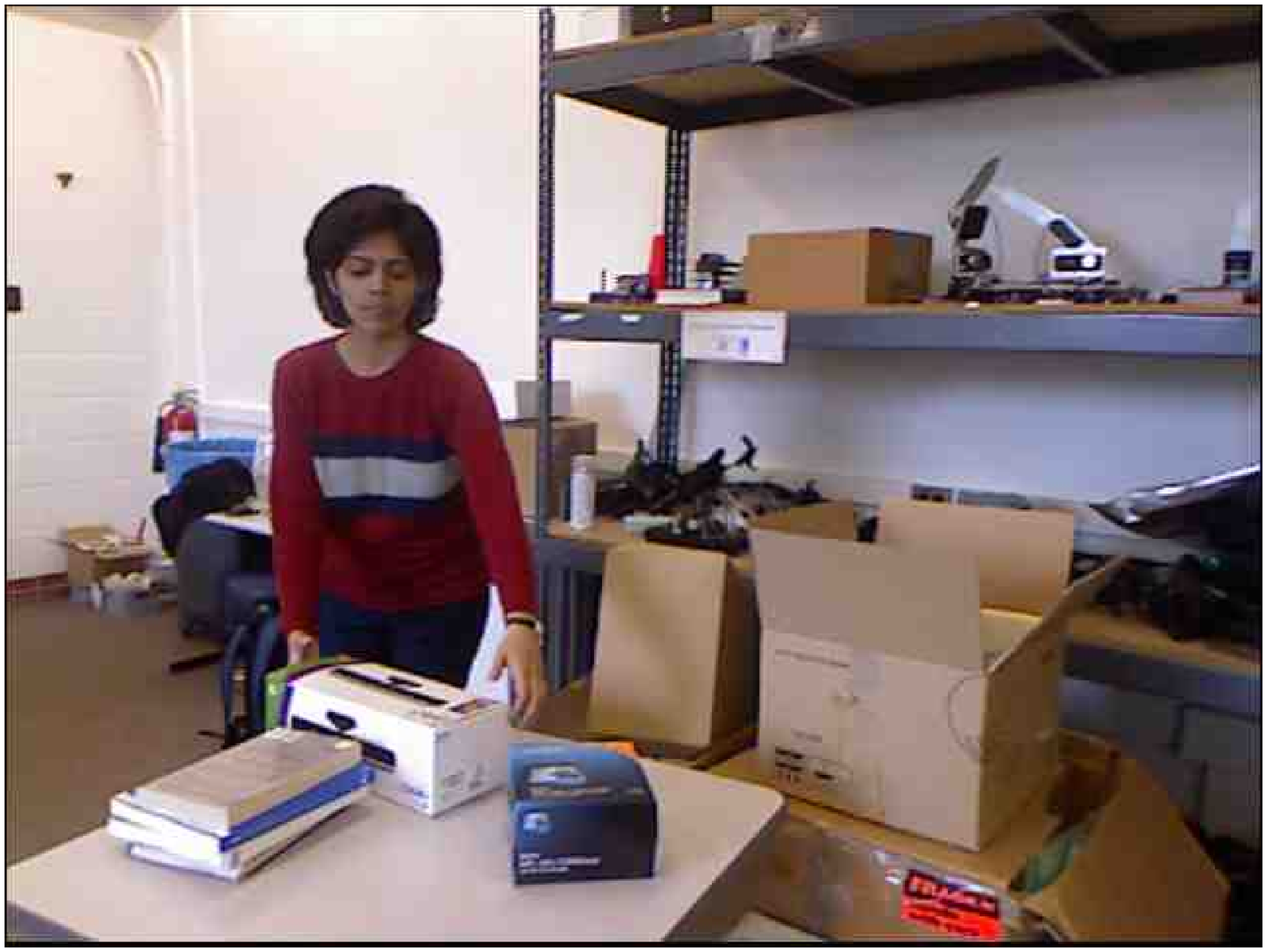}
\includegraphics[width=.16\linewidth,height=0.6in]{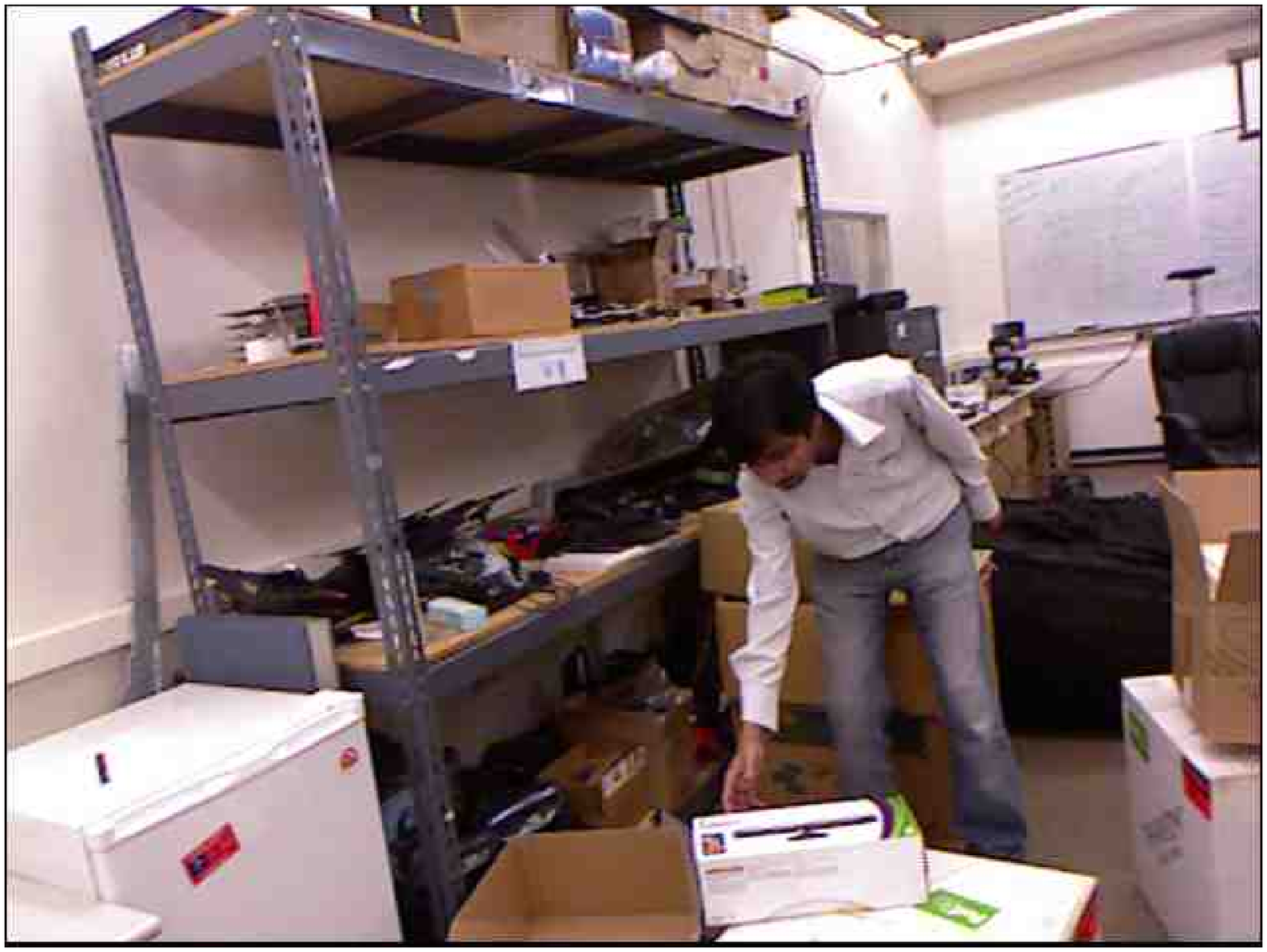}
\includegraphics[width=.16\linewidth,height=0.6in]{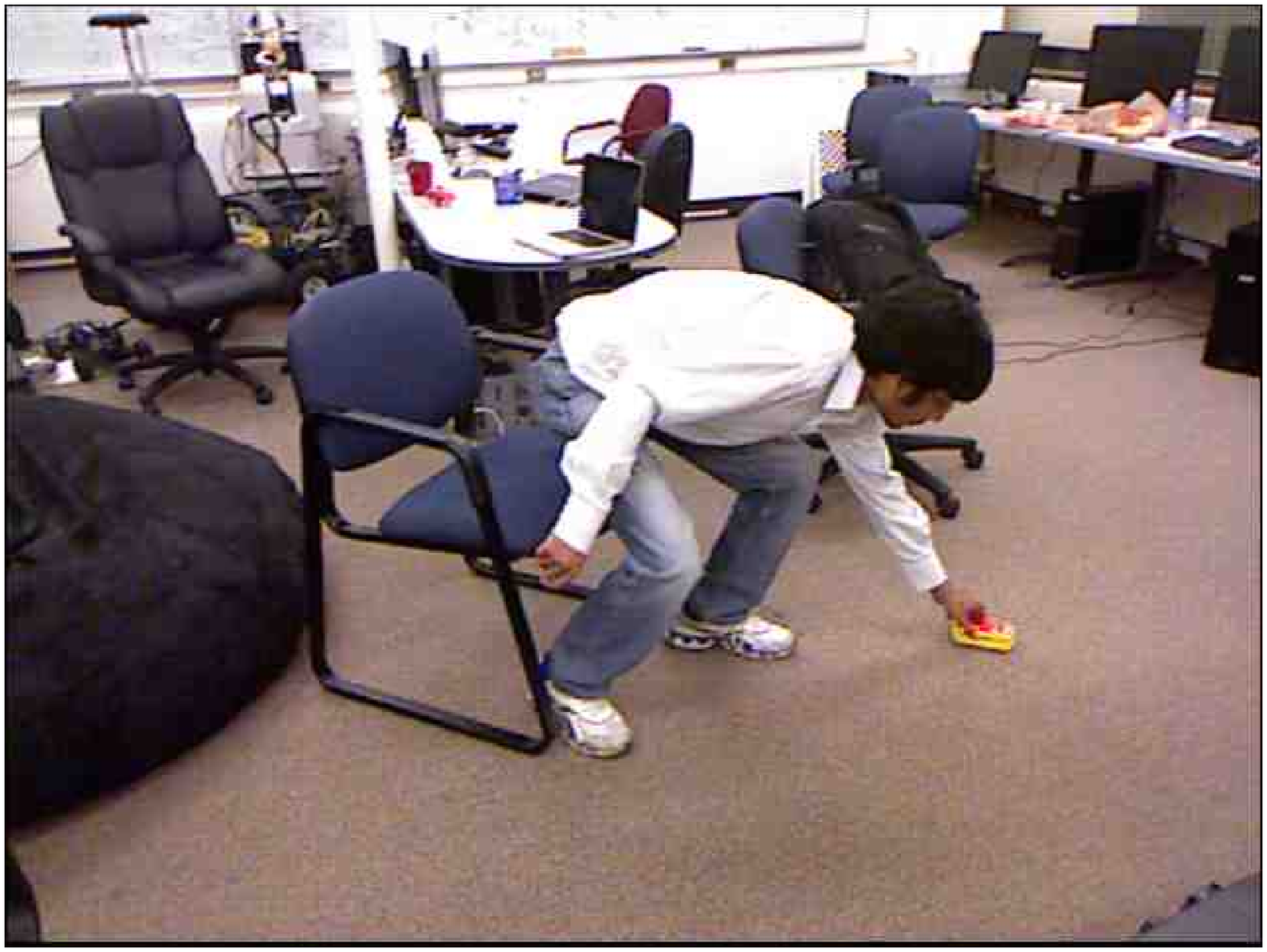}
\includegraphics[width=.16\linewidth,height=0.6in]{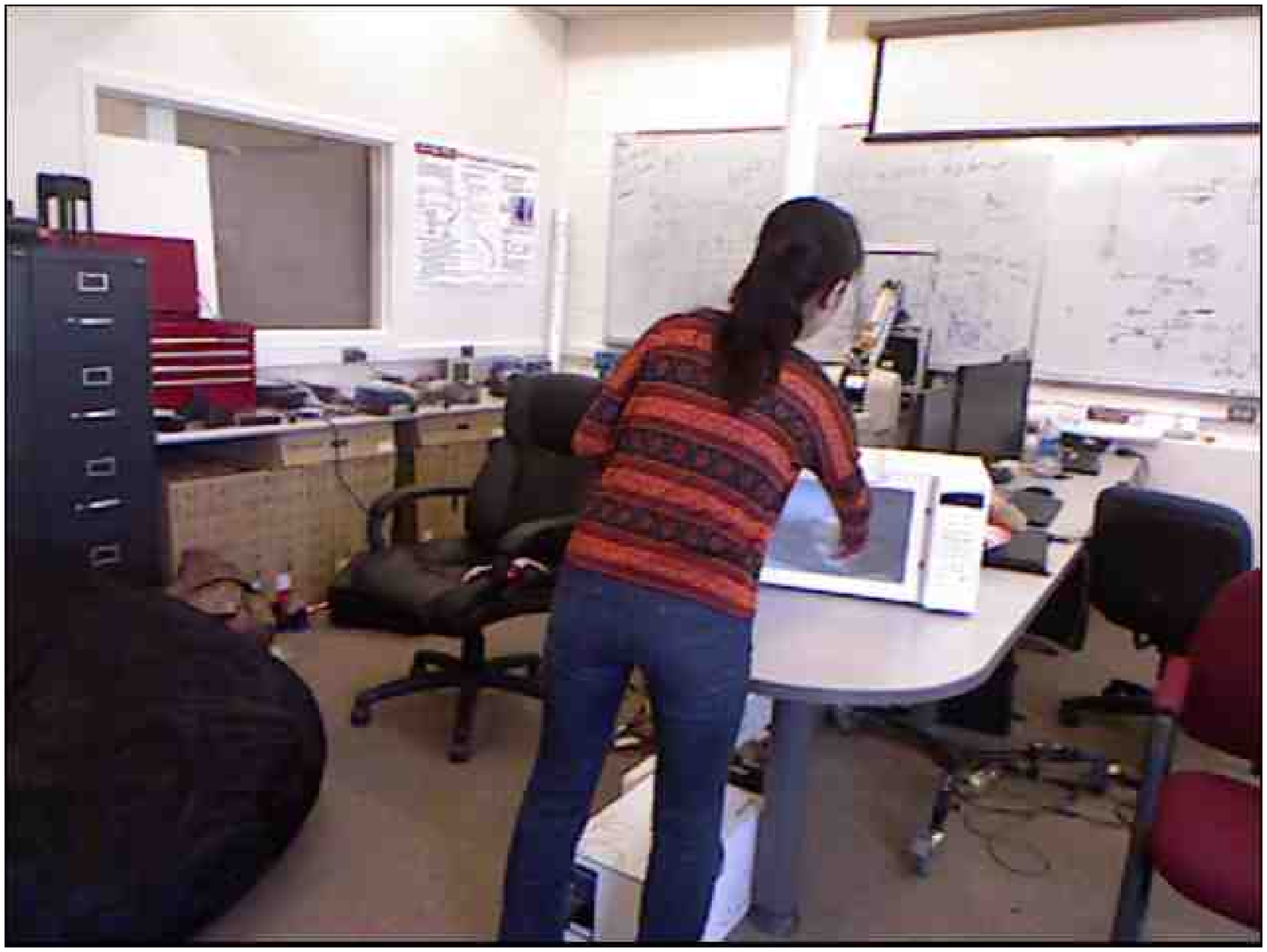}
\includegraphics[width=.16\linewidth,height=0.6in]{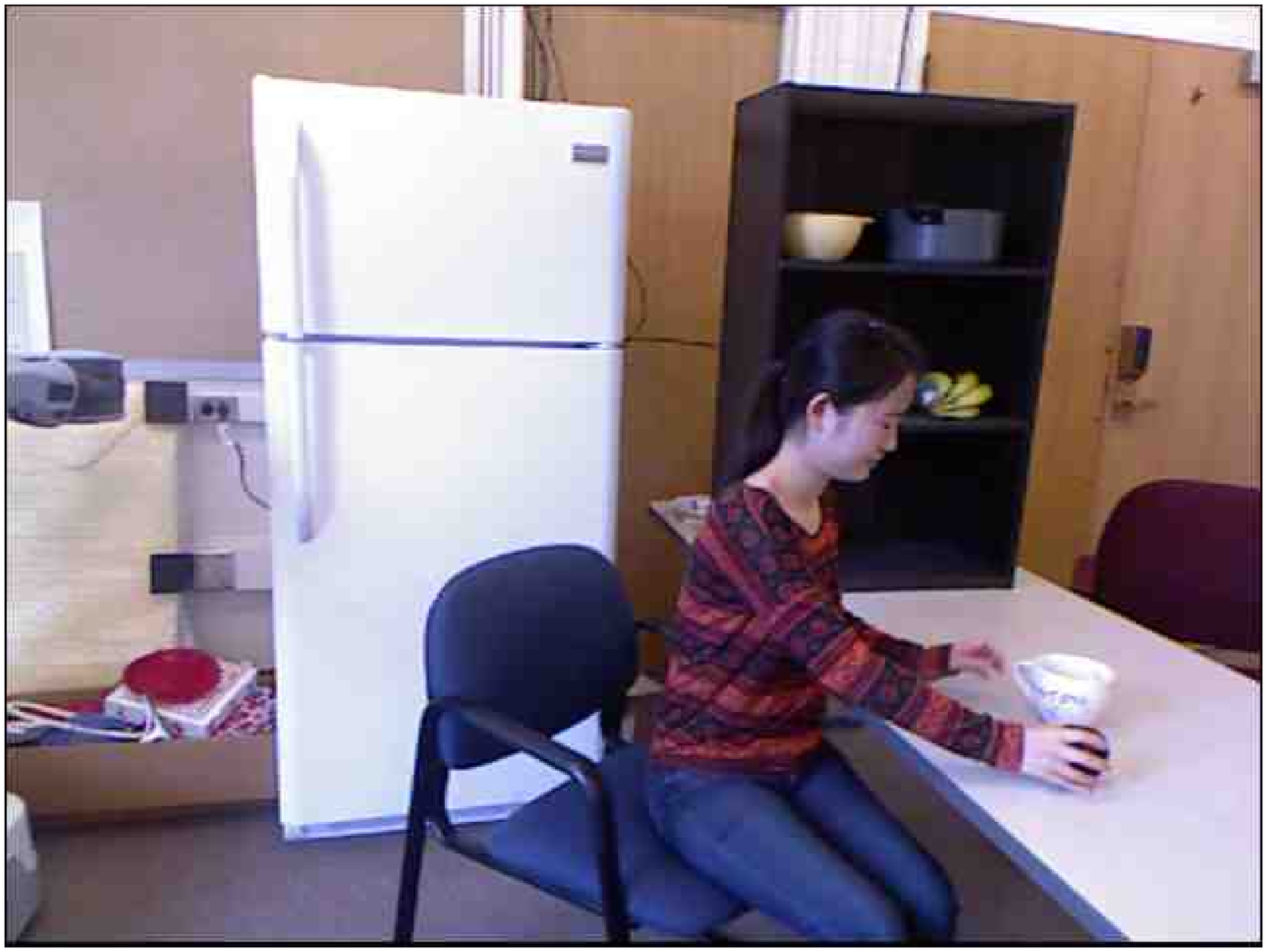}
\includegraphics[width=.16\linewidth,height=0.6in]{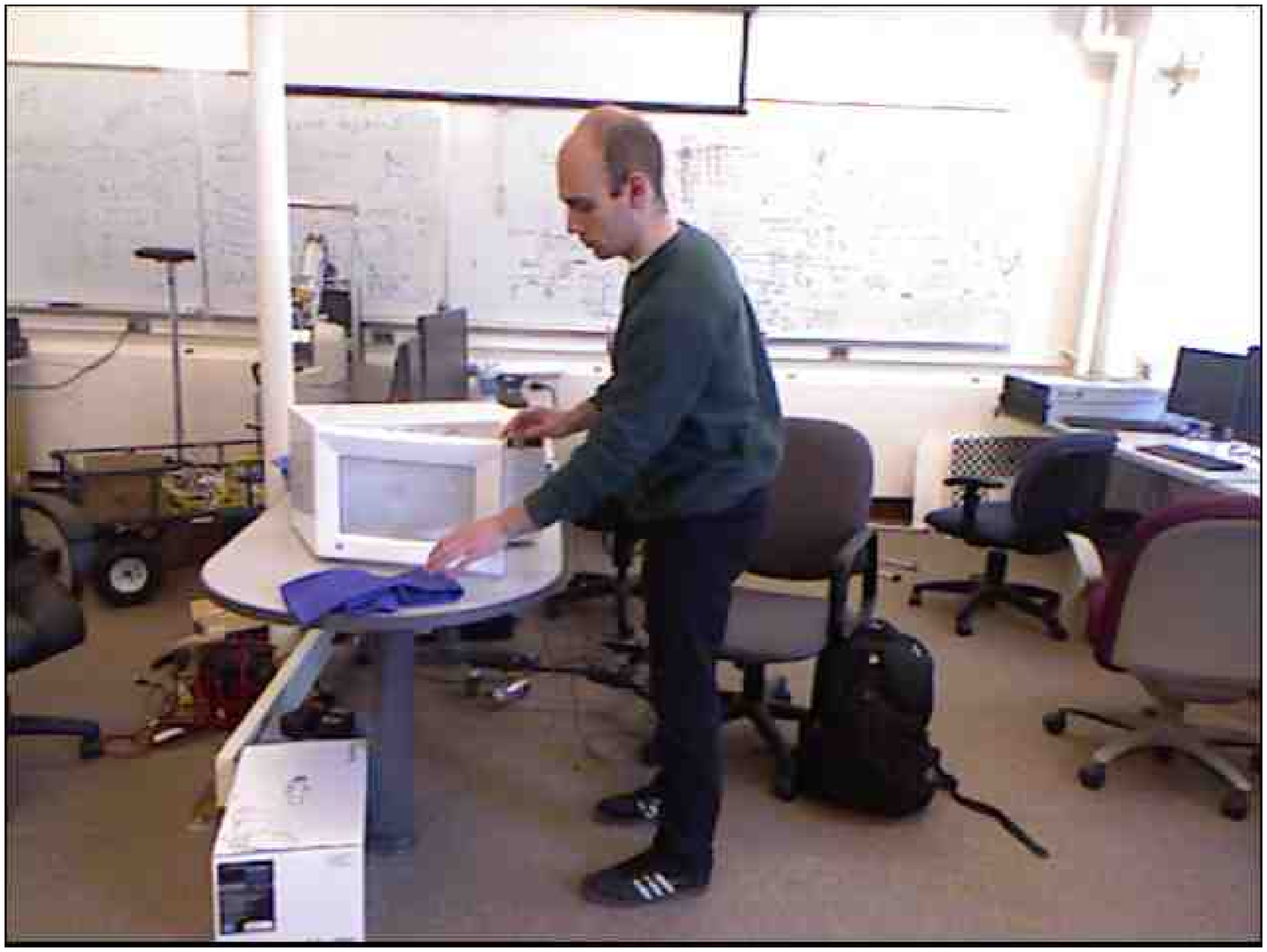}
\\
\includegraphics[width=.16\linewidth,height=0.6in]{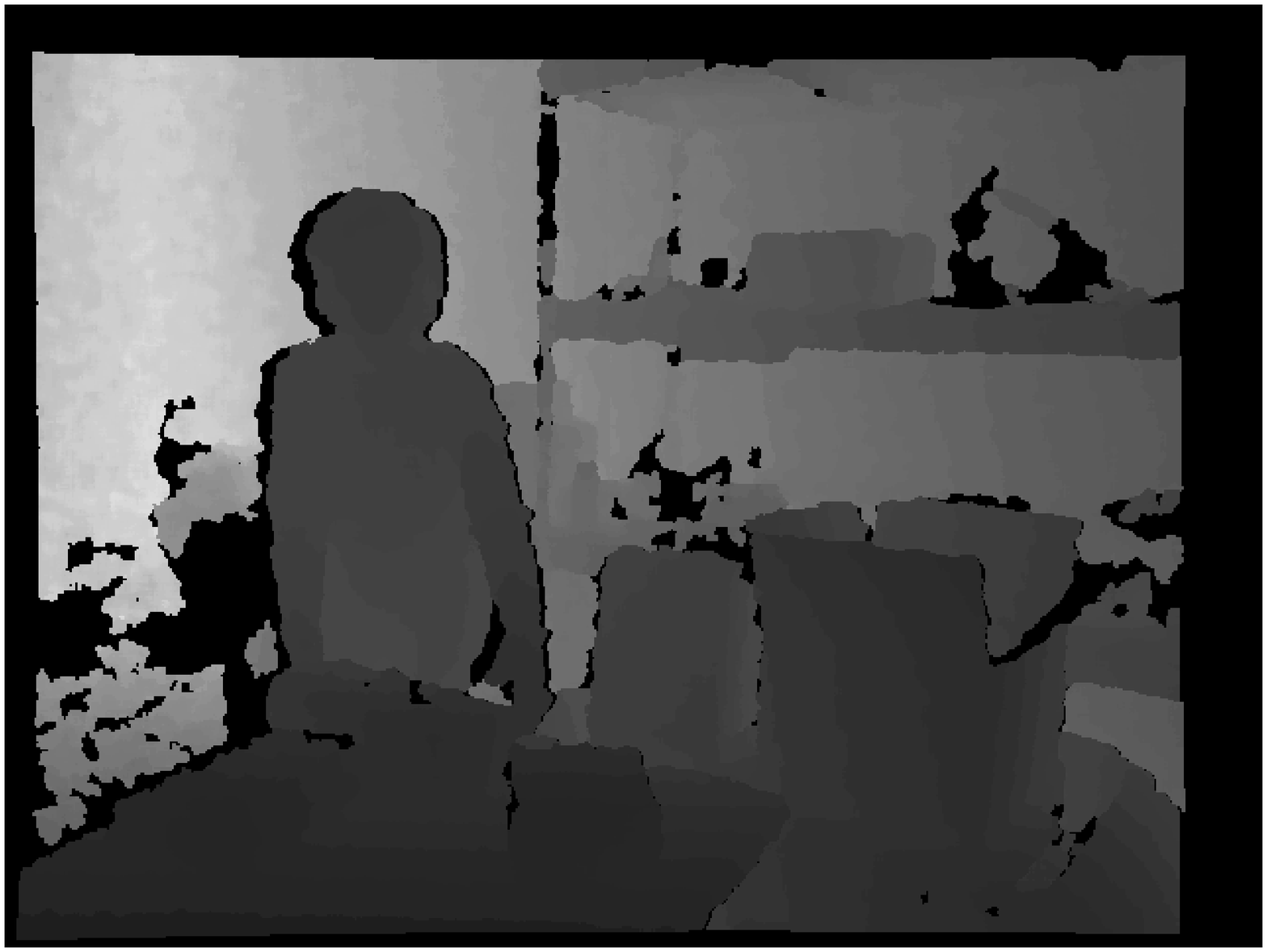}
\includegraphics[width=.16\linewidth,height=0.6in]{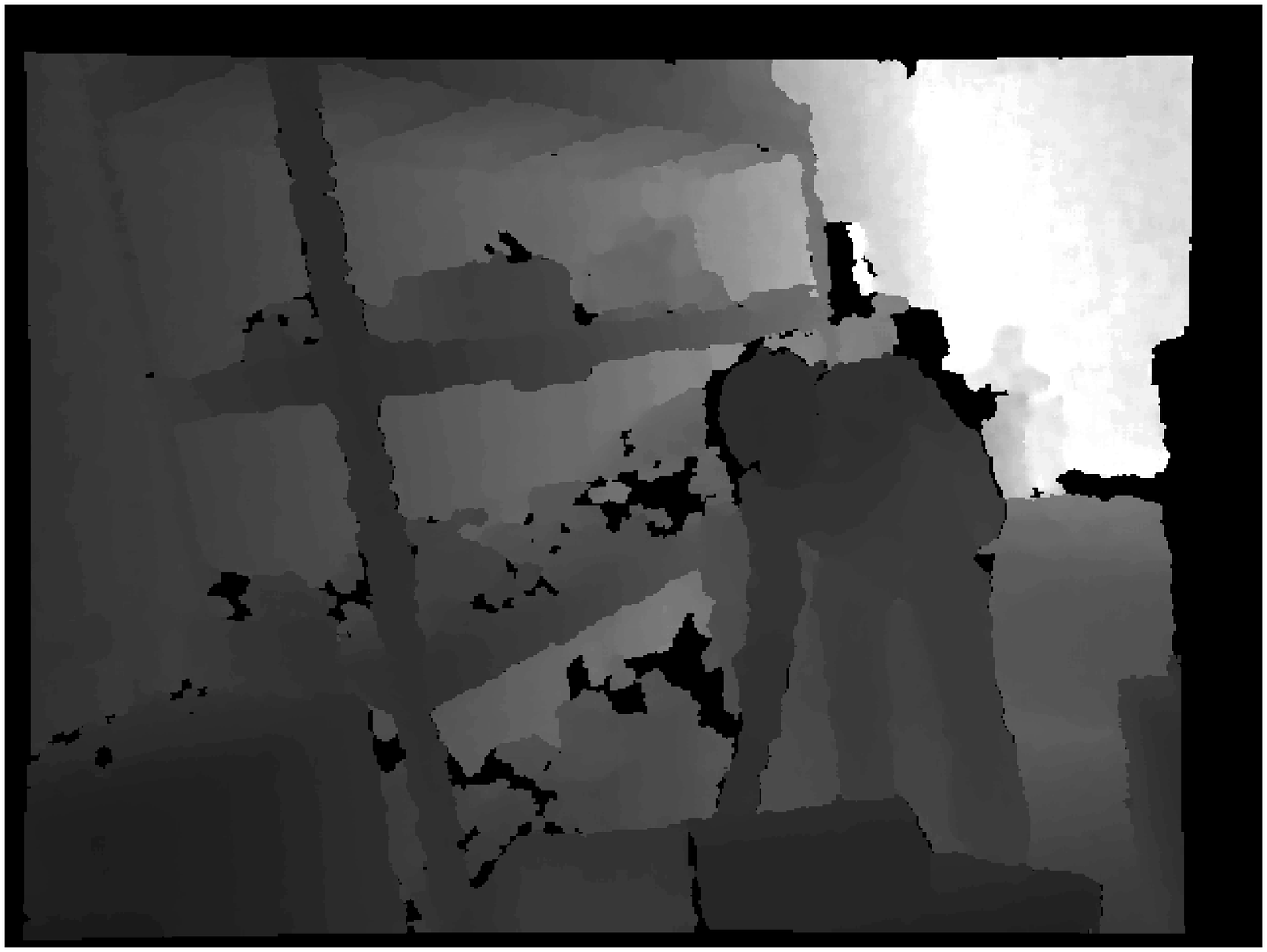}
\includegraphics[width=.16\linewidth,height=0.6in]{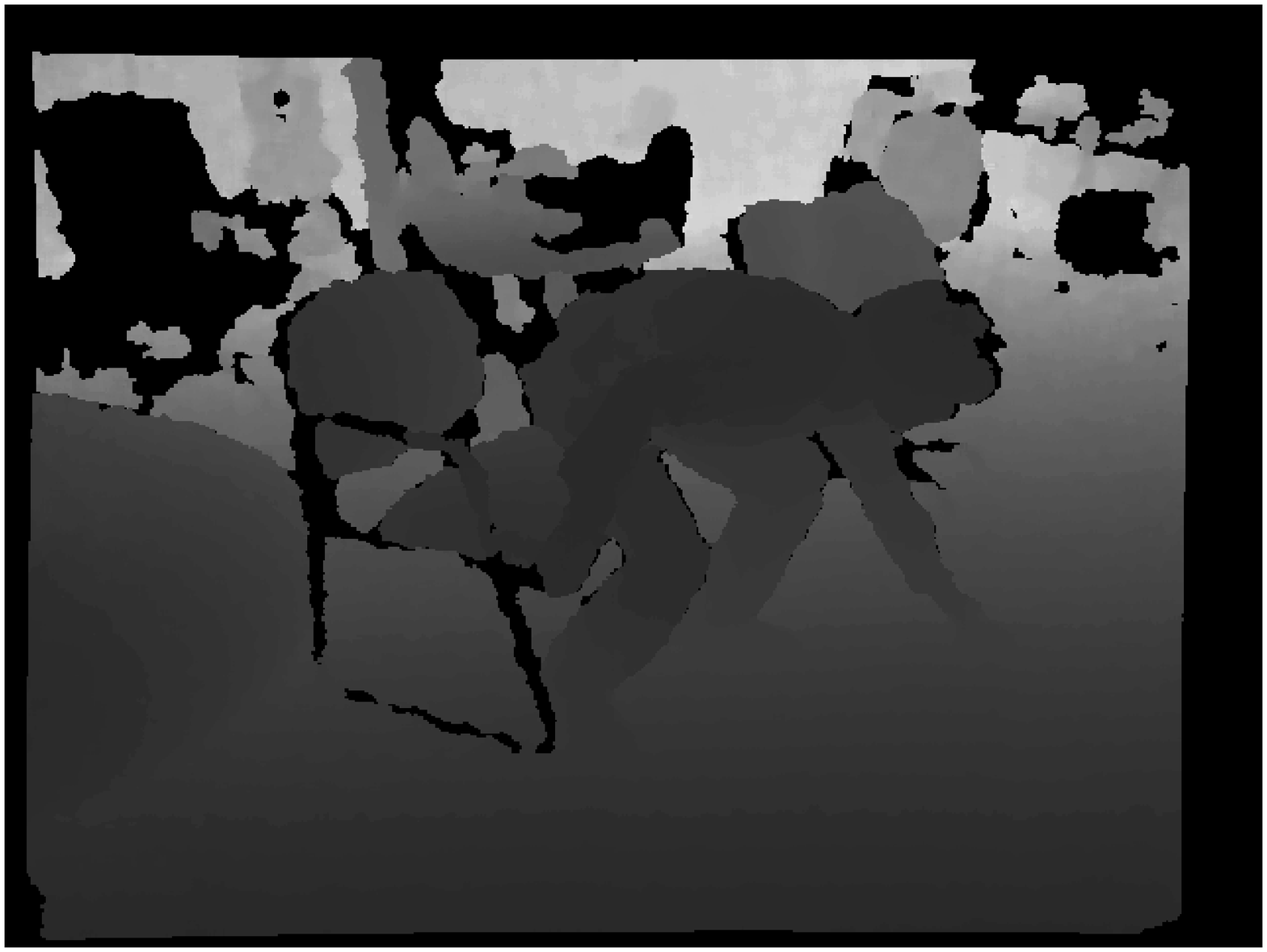}
\includegraphics[width=.16\linewidth,height=0.6in]{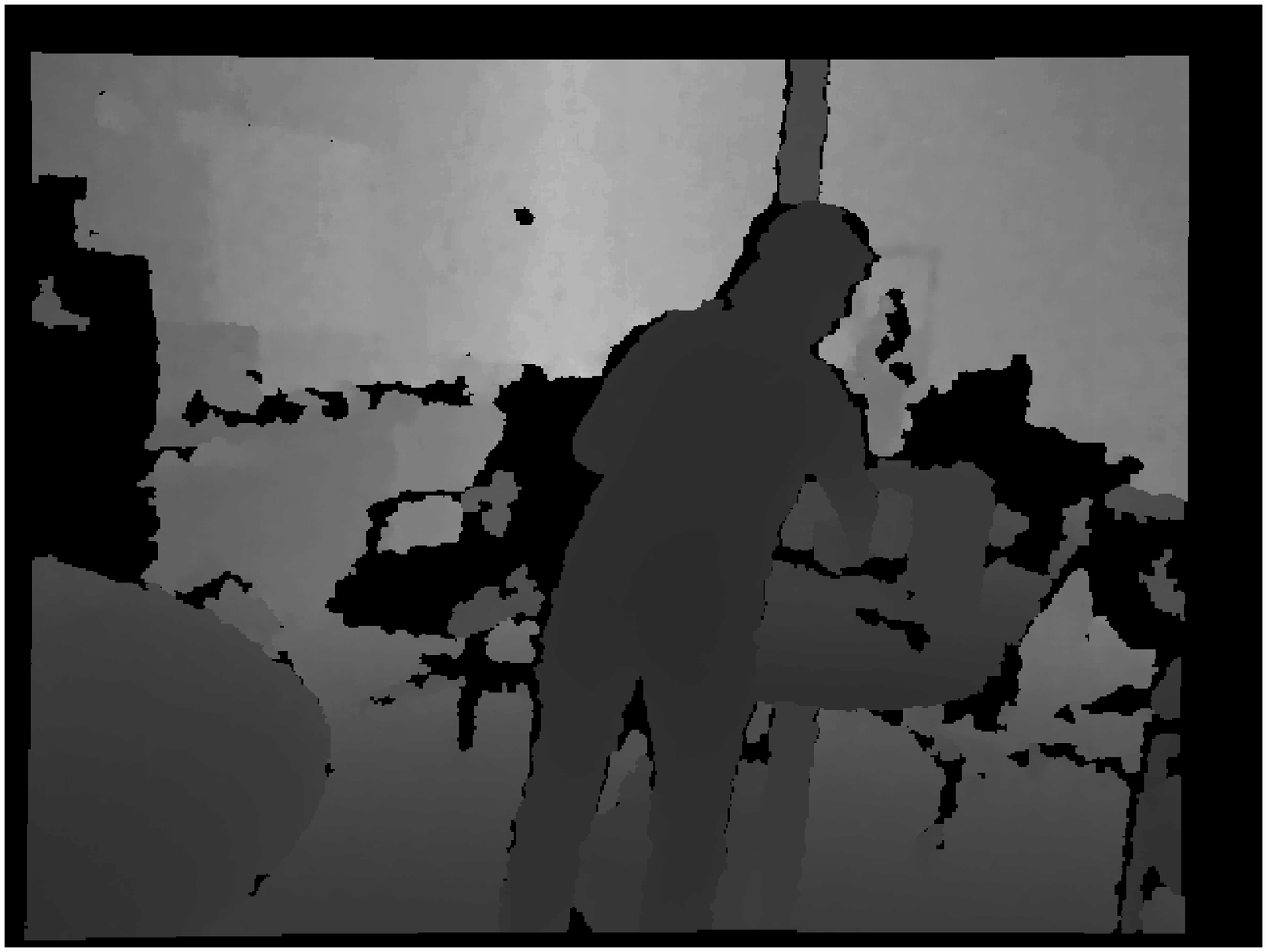}
\includegraphics[width=.16\linewidth,height=0.6in]{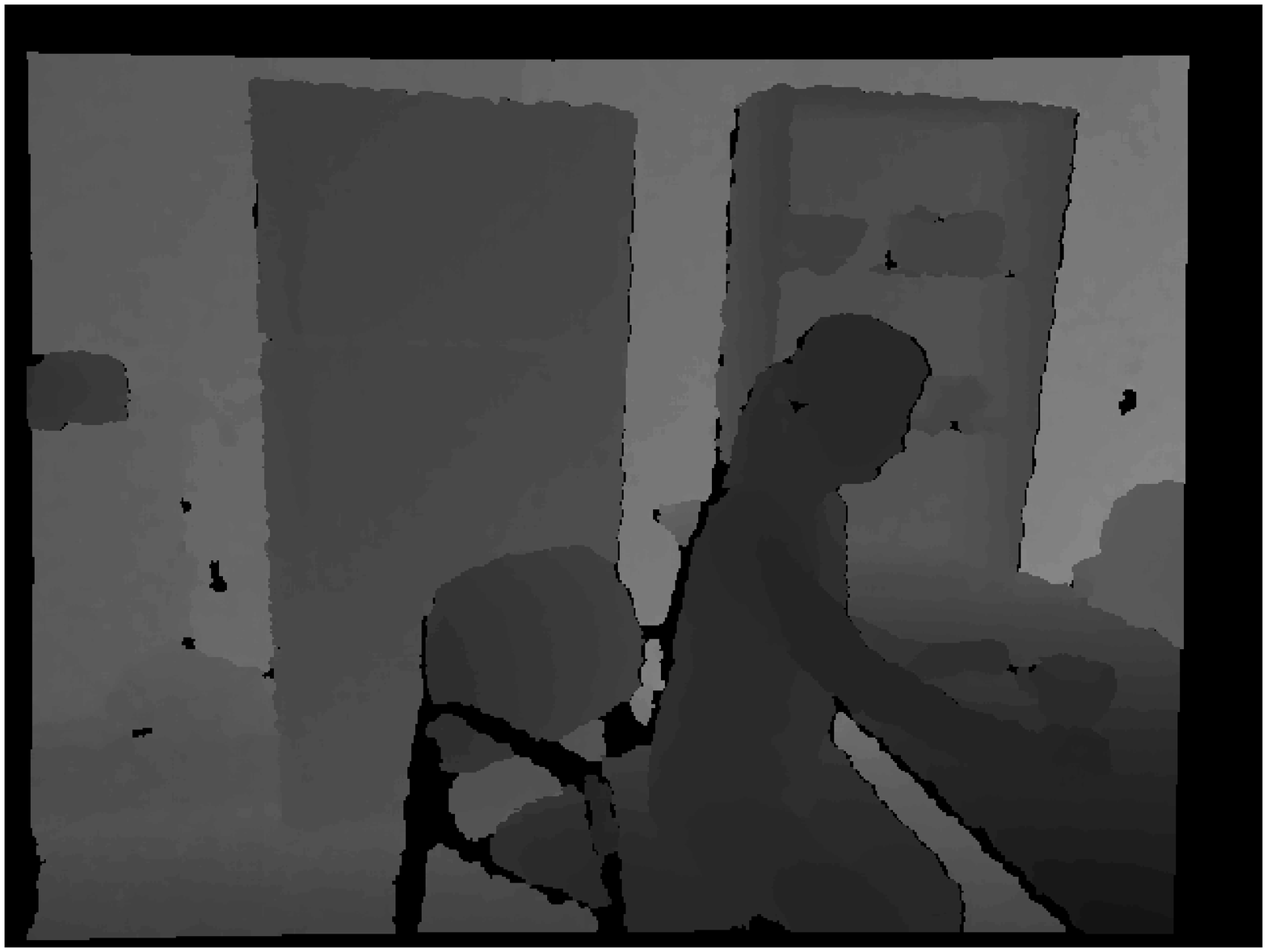}
\includegraphics[width=.16\linewidth,height=0.6in]{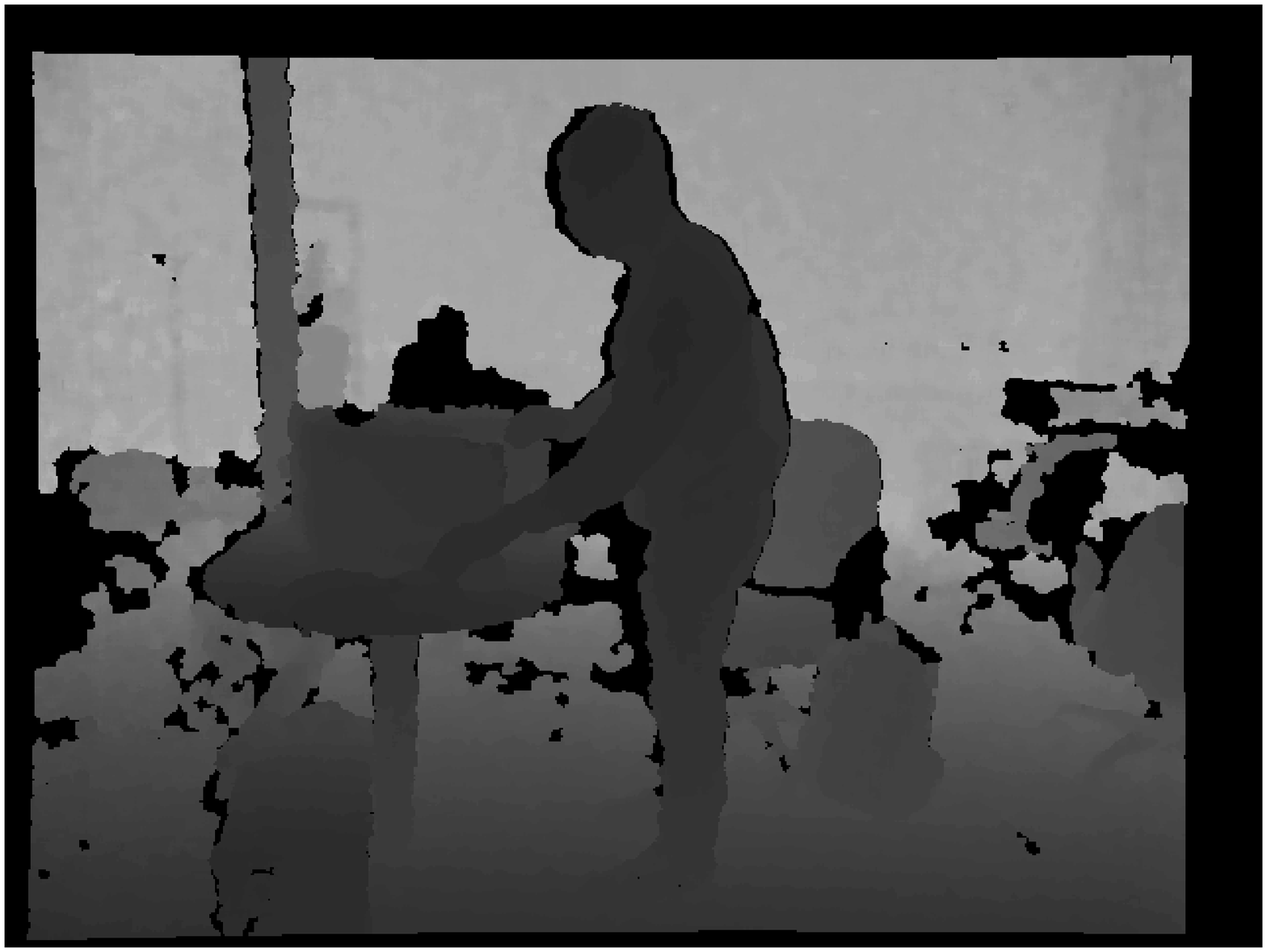}
 \vspace*{\captionReduceTop}
\caption{Example shots of \emph{reaching} sub-activity from our dataset. Top row shows
the RGB image, and the bottom row shows the corresponding depth images from the RGB-D camera.
Note that there are significant variations in the way the subjects perform the sub-activity. In addition, 
there is significant background clutter and subjects are partially occluded (e.g., column 1) or not facing the camera (e.g., column 4) in many instances.
}
\vskip -.17in
 \vspace*{\captionReduceBot}
 \label{fig:reachingData}
\end{figure}

Our focus in this work is to recognize complex human activities that take
 place over long time scales and that consist of a long sequence of sub-activities, 
 such as \emph{making cereal} and \emph{arranging objects} in a room. 
For example, \emph{making cereal} activity consists of around 12 sub-activities on average, which includes
\emph{reaching} the pitcher, \emph{moving} the pitcher to the bowl, and then \emph{pouring} the milk into the bowl. 
 This proves to be a very challenging task given the variability across individuals in performing each 
  sub-activity, and other environment induced conditions such as
  cluttered background and viewpoint changes. (See Fig.~\ref{fig:reachingData} for some examples.)
  
   In most previous works, object detection and activity recognition have been addressed as separate tasks. Only recently, some works have shown that modeling mutual context  is beneficial 
\cite{Gupta:TPAMI2009,Yao:CVPR10}. 
 The key idea in our work is to note that, in activity detection, it is sometimes more informative to 
 know \emph{how} an object is being used 
 (associated affordances, \cite{Gibson:1979}) rather than knowing \emph{what} the object is 
 (i.e., the object category). For example, both 
 chair and sofa might be categorized as `sittable,' and a cup might be categorized as both 
  `drinkable' and `pourable.'
Note that the affordances of an object change over time depending on its use, e.g.,
a pitcher may first be \emph{reachable}, then \emph{movable} and finally \emph{pourable}. 
In addition to helping activity recognition, recognizing object affordances is important by itself
because of their use in robotic applications (e.g.,  \cite{pancake_robot}).


We propose a method to learn human activities by modeling the sub-activities and affordances of the objects, 
how they change over time, and how they relate to each other.
More formally, we define a Markov Random Field
 over two kinds of nodes: object and sub-activity nodes. The edges in the graph  model the pairwise relations among interacting nodes, namely the object-object interactions, object-sub-activity interaction, and the temporal interactions. This model is built with each
spatio-temporal segment being a node. The parameters of this model are learnt using a structural SVM formulation \cite{Finley/Joachims/08a}. Given a new sequence of frames, we label the high-level activity, all the sub-activities and the object affordances using our learned model.

 However, the activities take place over a long time-scale, and different people execute sub-activities
 differently and for different periods of time. Furthermore, people also often merge two 
consecutive sub-activities together. 
Thus, segmentations in time are noisy and in fact, there may not be one `correct' segmentation, especially
at the boundaries. 
 One approach could be to consider \textit{all} possible segmentations,
 and marginalize the segmentation; however, this is computationally infeasible. In this work, we 
perform sampling of several segmentations, and consider labelings over these temporal segments
as latent variables in our learning algorithm. 




In extensive experiments over 120 activity videos collected from four subjects, 
we showed that our approach outperforms the baselines in both the tasks of activity as well as affordance detection. We achieved an accuracy of 91.8\% for affordance, 86.0\% for sub-activity labeling and 84.7\% for high-level activities respectively 
when given the ground truth segmentation, and an end-to-end accuracy of 83.9\%, 68.2\% and 
80.6\% on these respective tasks.


 
 

 


\vspace*{\sectionReduceTop}
\section{Related Work}\label{sec:relatedwork}
\vspace*{\sectionReduceBot}
There has been a lot of work on human activity detection from images \cite{YangWM10,Yao:ICCV11} 
and from videos \cite{action-jingen-wild,action-Laptev-cvpr08,actionbank,ShiIJCV11,HoaiLD11,model-recommendation-activity,complex-event-detection,fine-grained-activity,firstpersonactivity}. 
Here, we discuss works that are closely related to ours, and 
refer the reader to \cite{survey} for a survey of the field.
Most works (e.g., \cite{ShiIJCV11,HoaiLD11,model-recommendation-activity}) consider detecting actions at a `sub-activity' level (e.g., \emph{walk}, \emph{bend}, and \emph{draw}) instead of considering high-level activities. Their methods
range from discriminative learning techniques for joint segmentation and recognition \cite{ShiIJCV11,HoaiLD11} to combining multiple models \cite{model-recommendation-activity}.
Some works such as \cite{complex-event-detection} consider high-level activities. Tang et.~al.~\cite{complex-event-detection} propose a latent model for high-level activity classification and have the advantage of requiring only high-level activity labels for learning. 
None of these methods explicitly consider the role of objects or object affordances that not only
help in identifying sub-activities and high-level activities, but are also 
important for several robotic applications (e.g., \cite{pancake_robot}). 
    
  Some recent works \cite{Gupta:TPAMI2009,Yao:CVPR10,conf/icra/AksoyAWD10,tcsvt:actioncontext,firstpersonactivity} show that modeling the interaction 
 between human poses and objects in 2D videos 
results in a better performance on the tasks of object detection and activity recognition. 
However, these works cannot capture the rich 3D relations between the activities and objects, and 
are also fundamentally limited by the quality of the human pose inferred from the 2D data.
More importantly, for activity recognition, the object \emph{affordance} matters more than
its category.


Kjellstrom et.~al.~\cite{Kjellstrom:2011} 
used a Factorial CRF to 
simultaneously segment and classify human hand actions, as well as classify the object affordances involved in the activity from 2D videos. However, this work is limited to classifying only hand actions and does not model interactions between the objects. We consider complex full-body activities and show that modeling object-object interactions is important as objects have affordances even if they are not directly interacted with human hands. 



Recently, with the availability of inexpensive RGB-D sensors, some works consider
 labeling and object recognition in 3D point-clouds\cite{koppula:Nips11,lai:icra11b}.
Sung et.~al.~\cite{SungICRA2012} considered activity recognition from RGB-D videos. They propose a hierarchical maximum entropy Markov model to detect activities from RGB-D videos and treat the sub-activities as hidden nodes in their model. 
However, they use only human pose information for detecting activities and also constrain the number of sub-activities in each activity. In contrast, we model context from object interactions along with human pose,
also 
 present a better learning algorithm. (See Section~\ref{sec:experiments} for further comparisons.)
Gall et.~al.~\cite{GallCVPR11} also use depth data to perform sub-activity (referred to as action) classification and functional categorization of objects. 
Their method first detects the sub-activity being performed using the estimated human pose from depth data, and then 
performs object localization and clustering of the objects into functional categories based on the detected sub-activity. 
In contrast, our proposed method performs joint sub-activity and affordance labeling and uses these labels to 
perform high-level activity detection.

All above works lack a unified framework of combining all the information available in human interaction activities and therefore we propose a model that  captures both the spatial and temporal relations between objects and human poses to perform joint object affordance and activity detection.

\vspace*{\sectionReduceTop}
\section{Our Approach}
\vspace*{\sectionReduceBot}

\begin{figure}[tb!]
 \centering
 \vskip -.15in
 \includegraphics[width=.95\linewidth, height=1.5in]{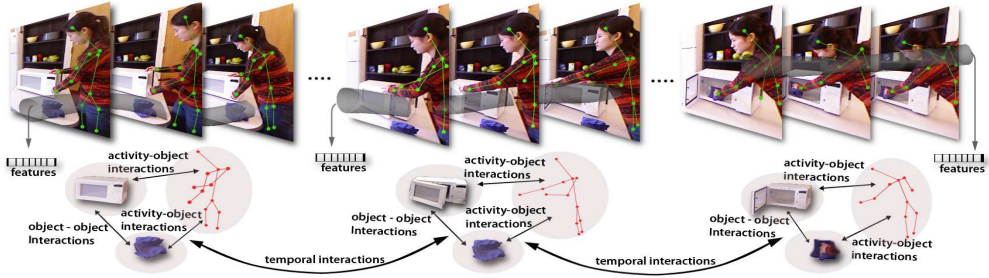}
 \vspace*{\captionReduceTop}
 \caption{Pictorial representation of the different types of nodes and relationships modeled in part of the \emph{cleaning objects} activity comprising three sub-activities: \emph{reaching}, \emph{opening} and \emph{scrubbing}.}
 \vspace*{\captionReduceBot}
\vskip -.05in
 \label{fig:modelfigure}
 \end{figure}

Our goal is to perform joint activity and object affordance labeling of RGBD videos.
As illustrated in Fig.~\ref{fig:modelfigure}, we define a Markov Random Field (MRF)
over the spatio-temporal sequence we get from an RGBD video. The MRF 
is represented as a graph $\mathcal{G} = (\mathcal{V},\mathcal{E})$. 
There are two types of nodes in  $\mathcal{G}$: objects nodes denoted by $\mathcal{V}_o$ and 
sub-activity nodes denoted by $\mathcal{V}_a$.


\vspace*{\subsectionReduceTop}
\subsection{Representation}
\vspace*{\subsectionReduceBot}

If we build our graph with nodes for objects and sub-activities for each time instant (at 30fps), then we
will end up with quite a large graph. Furthermore, such a graph would not be able to model
meaningful transitions between the sub-activities because they take place over a long-time (e.g., a few seconds). 
Therefore, in our approach we first segment the video into small temporal segments, and our goal
is to label each segment with appropriate labels. 
We try to over-segment, so that we end up
with more segments and avoid merging two sub-activities into one segment.
Each of these segments occupies a small length of time and therefore, considering nodes per segment 
gives us a meaningful and concise representation for the graph $\mathcal{G}$.  
With such a representation, we can model meaningful transitions of a sub-activity following another, e.g., \emph{pouring}  followed by \emph{moving}. 



\vspace*{\subsectionReduceTop}
\subsection{Overview of Properties Captured\label{sec:relations}}
\vspace*{\subsectionReduceBot}

Over the course of a video, a human may interact with several objects and perform
several sub-activities over time. In our MRF model (in Section~\ref{sec:model}), these interactions
are represented by the edges $\mathcal{E}$, and we try
to capture the following properties:

{\bf Affordance - sub-activity relations.} At any given time, the affordance of the 
object depends on the sub-activity it is involved in. For example, a cup has the affordance 
of \emph{`pourable'} in a \emph{pouring} sub-action and has the affordance of \emph{`drinkable'} 
in a \emph{drinking} sub-action. 
We compute relative geometric features between the object and 
the human's skeletal joints to capture this.

{\bf Affordance - affordance relations.}  
 Objects have affordances even if they are not interacted directly with by the human,
 and their affordances depend on the affordances of other objects around them.
E.g., in the case of \emph{pouring} from a pitcher to a cup, the cup is not 
interacted by the human directly but has the affordance \emph{`pour-to'}. 
We therefore use relative geometric features such as ``on top of", ``nearby", ``in front of", etc., 
to model the affordance - affordance relations.

{\bf Sub-activity change over time.} Each activity consists of a sequence of sub-activities that
change over the course of performing the activity. We model this by incorporating 
temporal edges in $\mathcal{G}$.

{\bf Affordance change over time.} The object affordances depend on the activity being 
performed and hence change along with the sub-activity over time. We model the temporal 
change in affordances of each object using features such as change in appearance 
and location of the object over time.

\vspace*{\subsectionReduceTop}
\subsection{Model}
\vspace*{\subsectionReduceBot}
\label{sec:model}
We model the spatio-temporal structure of an activity using a model isomorphic to a Markov Random Field with log-linear node and pairwise edge potentials (see Fig.~\ref{fig:modelfigure} 
for an illustration). 
Let $K_a$ denote the set of sub-activity labels, and $K_o$ denote the set of object affordance labels.  
Given a temporally segmented 
3D video $\x=(\xs{1},...,\xs{N})$ consisting of temporal segments $\xs{s}$, we aim to predict a labeling $\y=(\ys{1},...,\ys{N})$ for each segment.
For a segmented 3D video $\x$, the prediction $\hat{\y}$ is computed as the argmax of a discriminant function $\df{\x}{\y}{\w}$ that is parameterized by weights $\w$.
{\small
\begin{equation} 
\hat{\y} = \argmax_\y \df{\x}{\y}{\w}
\label{eq:argmax}
\end{equation}   
}
\vskip -.2in
The discriminant function captures the dependencies between the sub-activity and object affordance labels as defined by an undirected graph $\mathcal{G}  = (\mathcal{V},\mathcal{E})$. 
We now describe the structure of this graph. 
For object nodes denoted by $\mathcal{V}_o$ and sub-activity nodes denoted by $\mathcal{V}_a$,
let $\mathcal{V}_o^s$ denote set of object nodes of segment $s$, and $v_a^s$ denote the sub-activity node of segment $s$. For all segments $s$,  there is an edge connecting all the nodes in $\mathcal{V}_o^s$ to each other and to the sub-activity node  $v_a^s$. These edges signify the relationships within the objects, and between the objects and the human pose within a segment and are referred to as \emph{`object - object interactions'} and 
\emph{`sub-activity - object interactions'}  in the Fig.~\ref{fig:modelfigure} respectively. 

The sub-activity node of segment $s$ is connected to the sub-activity nodes in segments $(s-1)$ and 
$(s+1)$. Similarly every object node of segment $s$ is connected to the corresponding object nodes in segments $(s-1)$ and 
$(s+1)$. These edges model the \emph{temporal interactions} between the human poses and the objects respectively and represented by doted edges in the Fig.~\ref{fig:modelfigure}. 

We categorize the edges into different types denoted by $\cal T$. The various types are \emph{object - object} edges and \emph{object - sub-activity} edges and \emph{sub-activity - sub-activity} edges. Let $\mathcal{E}_t$ denote the edges of type $t \in \cal T$ and $ T_t $ denote pairs of possible labels for the nodes connected by edges of type $t$. For example if the edge type $t$ is object - sub-activity, then $(l,k) \in T_t, \forall l \in K_o,  \forall k \in K_a $. 



Let $\ysc{i}{k}$ be a binary variable representing the node $i$ having label $k$, where $k \in K_o$ for object nodes and $k \in K_a$ for sub-activity nodes. All $k$ binary variables together represent the label of a node in the above described graph $\mathcal{G}$. A segment label $\ys{s}$ is composed of two types of node labels: a set of object affordance labels, 
$\{ \ysc{i}{k} : k \in K_o; \forall i \in  \mathcal{V}_o^s\}$, and a sub-activity label  $\{\ysc{i}{k}: k \in K_a; i = v_a^s\}$.
 Given $\mathcal{G}$, we define the following discriminant function based on individual node features $\fs{i}$ and $\fo{i}$ and edge features $\fe{t}{i}{j}$ as further described below.
\vskip -.15in
{\small
\begin{eqnarray} 
\df{\y}{\x}{\w} & = &  \sum_{i \in \mathcal{V}_a}  \sum_{k \in K_a} \ysc{i}{k} \left[\ws{k} \cdot \fs{i} \right] + 
 \sum_{i \in \mathcal{V}_o} \sum_{k \in K_o}  \ysc{i}{k} \left[\wo{k} \cdot \fo{i} \right] \nonumber \\
& & +  \sum_{t \in {\cal T}} \sum_{(i,j)\in \mathcal{E}_t}   \sum_{(l,k)\in T_t} \ysc{i}{l} \ysc{j}{k}  \left[\we{t}{l}{k} \cdot \fe{t}{i}{j}\right]  
\label{eq:model}
\end{eqnarray}
}
\vskip -.1in
For the node feature maps, $\fs{i}$ and $\fo{i}$, there is one weight vector for each of the sub-activity classes in $K_a$ and each of the object affordance classes in $K_o$ respectively. There are multiple types $t$ of the edge feature maps $\fe{t}{i}{j}$, 
each corresponding to a different type of edge in the graph.
For each type, there is one weight vector each for every pair of labels the edge can take.

 \textbf{Features.}  
For a given object node $i$, the node feature map $\fo{i}$ is a vector of features representing the object's location in the scene and how it changes within the temporal segment. These features include the $(x,y,z)$ coordinates of the object's centroid and the coordinates of the object's bounding box at the middle frame of the temporal segment. We also run a SIFT feature based object tracker \cite{PeleECCV2008} to find the corresponding 
points between the adjacent frames and then compute the transformation matrix based on the matched image points. We add the transformation matrix corresponding to the object in the middle frame with respect to its previous frame to the features in order to capture the object's motion information.
 In addition to the above features, we also compute the total displacement and the total distance moved by the object's centroid in the set of frames belonging to the temporal segment. We then perform cumulative binning of the feature values into 10 bins.
In our experiments, we have $\fo{i} \in \Re^{180}$.

Similarly, for a given sub-activity node $i$, the node feature map $\fs{i}$ gives a vector of features computed using the human skeleton information obtained from running Openni's skeleton 
tracker \cite{openni-tracker} on the RGBD video.  We compute the features described above 
for each the upper-skeleton joint (neck, torso, left shoulder, left elbow, left palm, right shoulder, right elbow and right palm) locations relative to the subject's head location, thus giving us $\fs{i} \in \Re^{1030}$.

 The edge feature maps $\fe{t}{i}{j}$ describe the relationship between node $i$ and $j$. 
For capturing the \emph{object-object} relations within a temporal segment,
we compute relative geometric features such as the difference in $(x,y,z)$ coordinates of the object centroids and the distance between them. These features are computed at the first, middle and last frames of the temporal segment along with $min$ and $max$ of their values across all frames in the temporal segment to capture the relative motion information. 
This gives us $\fe{1}{i}{j} \in \Re^{200}$. Similarly for \emph{object--sub-activity} relation features $\fe{2}{i}{j} \in \Re^{400}$, we use the same features as for the \emph{object-object} relation features, but we compute them between the upper-skeleton joints and the each object's centroid. The temporal relational features capture the change across temporal segments and we use the vertical change in position and the distance between by corresponding object and the joint locations. 
This
gives us $\fe{3}{i}{j} \in \Re^{40}$ and $\fe{4}{i}{j} \in \Re^{160}$ respectively.

\noindent
\textbf{Inference.}  Given the model parameters $w$, the inference problem is to find the best labeling $\hat{\y}$ for a new video $\x$, i.e., solving the argmax in Eq.~(\ref{eq:argmax}) for the discriminant function in Eq.~(\ref{eq:model}). This is a NP hard problem. However, its equivalent formulation as the following mixed-integer program has a linear relaxation
which can be solved efficiently as a quadratic pseudo-Boolean optimization problem using a graph-cut method
\cite{Kolmogorov/Rother/07}.
 \vskip -.2in
{\small 
\begin{align}
& \hat{\y}\!=\!\argmax_{\y}\max_{\mathbf z} \sum_{i \in \mathcal{V}_a}  \sum_{k \in K_a} \ysc{i}{k} \left[\ws{k} \cdot \fs{i} \right] + \!\!\! \sum_{i \in \mathcal{V}_o} \sum_{k \in K_o}  \ysc{i}{k} \left[\wo{k} \cdot \fo{i} \right] \nonumber \\
& \qquad \qquad \qquad  \quad  +  \sum_{t \in {\cal T}} \sum_{(i,j)\in \mathcal{E}_t}   \sum_{(l,k)\in T_t} \zsc{ij}{lk} \left[\we{t}{l}{k} \cdot \fe{t}{i}{j}\right]  
 \label{eq:relaxobj}\\
& \forall i,j,l,k\!: \zsc{ij}{lk}\le \ysc{i}{l}, \zsc{ij}{lk}\le \ysc{j}{k}, 
\ysc{i}{l} + \ysc{j}{k} \le \zsc{ij}{lk}+1,
\zsc{ij}{lk},\ysc{i}{l} \in \{ 0,1 \} 
\label{eq:relaxconst}
\end{align}
}
\vskip -.21in
Note that the products $\ysc{i}{l} \ysc{j}{k}$ have been replaced by auxiliary variables $z^{lk}_{ij}$. 
Relaxing the variables $\zsc{ij}{lk}$ and $\ysc{i}{l}$ to the interval $[0,1]$ results in a linear program that can be 
 shown to always have half-integral solutions (i.e., $\ysc{i}{l}$ only take values $\{0,0.5,1\}$ at the solution) \cite{hammer1984roof}. 
Since every node in our experiments has exactly one class label, we also consider the linear relaxation from above with the additional constraints $\forall i \in \mathcal{V}_a: \sum_{l \in K_a} \ysc{i}{l} = 1$ and $\forall i \in \mathcal{V}_o: \sum_{l \in K_o} \ysc{i}{l} = 1$. This problem can no longer be solved via graph cuts.
We compute the exact mixed integer solution including these additional constraint using a general-purpose MIP solver\footnote{http://www.tfinley.net/software/pyglpk/readme.html} during inference. 
The MIP solver takes 10.7 seconds on an average for one video (a typical video 
has a graph with 17 sub-activity nodes and 592 object nodes, i.e., 6090 variables).

\textbf{Learning.} We take a large-margin approach to learning the parameter vector $\w$ of Eq.~(\ref{eq:model}) from labeled training examples $(\x_1,\y_1),...,(\x_\M,\y_\M)$ 
\cite{Taskar/AMN,Tsochantaridis/04}.
Our method optimizes a regularized upper bound on the training error
$R(h) = \frac{1}{\M} \sum_{m=1}^{\M} \loss{\y_m}{\hat{\y}_m}$,
where $\hat{\y}_m$ is the optimal solution of Eq.~(\ref{eq:argmax}) and $\loss{\y}{\hat{\y}}=\sum_{i \in \mathcal{V}_o} \sum_{k \in K_o} |\ysc{i}{k} - \hat{\ysc{i}{k}}| + \:\:\:\sum_{i \in \mathcal{V}_a} \sum_{k \in K_a} |\ysc{i}{k} - \hat{\ysc{i}{k}}|$. To simplify notation, note that Eq.~(\ref{eq:relaxobj}) can be equivalently written as $\w^T \Psi(\x,\y)$ by appropriately stacking the $\ws{k}$ , $\wo{k}$ and $\we{t}{l}{k}$ into $\w$ and the $\ysc{i}{k}\fs{i}$, $\ysc{i}{k}\fo{i}$ and $\zsc{ij}{lk}\fe{t}{i}{j}$ into $\Psi(\x,\y)$, where each $\zsc{ij}{lk}$ is consistent with Eq.~(\ref{eq:relaxconst}) given $\y$. Training can then be formulated as the following convex quadratic program \cite{joachims2009cutting}:
{\small
\begin{eqnarray} \label{eq:trainqp}
\min_{w,\xi} & &  \frac{1}{2} \w^T\w + C\xi\\
s.t. & &   \forall \bar{\y}_1,...,\bar{\y}_\M \in \{0,0.5,1\}^{N \cdot K} :
 \frac{1}{M} \w^T \sum_{m=1}^{M} [\Psi( \x_m, \y_m) - \Psi(\x_m,\bar{\y}_m)] \ge \Delta(\y_m,\bar{\y}_m) -\xi \nonumber
\end{eqnarray}
}
\vskip -.1in
While the number of constraints in this QP is exponential in $\M$, $N$ and $K$, it can nevertheless be solved efficiently using the cutting-plane algorithm 
\cite{joachims2009cutting}. 
The algorithm needs access to an efficient method for computing
{\small
\begin{eqnarray}
\bar{\y}_m & = & \!\!\!\!\!\argmax_{\y \in \{0,0.5,1\}^{N \cdot K}} \left[ \w^T \Psi(\x_m,\y) + \loss{\y_m}{\y} \right].
\end{eqnarray} }
Due to the structure of $\loss{.}{.}$, this problem is identical to the relaxed prediction problem in Eqs.~(\ref{eq:relaxobj})-(\ref{eq:relaxconst}) and can be solved efficiently using graph cuts.



\vspace*{\subsectionReduceTop}
\subsection{Multiple Segmentations}
\label{sec:segmentation}
\vspace*{\subsectionReduceBot}

Segmenting an RGB-D video in time can be noisy, and multiple segmentations may be valid. 
Therefore, we will perform multiple segmentations by using different methods and criterion
of segmentation (see Section~\ref{sec:experiments} for details). Thus, we get a set 
$\mathcal{H}$ of multiple segmentations, and
 let $h_n$ be the $n^{th}$ segmentation.
A discriminant function $\df{\xh{n}}{\yh{n}}{\wh{n}}$ can now be defined for each 
$h_n$ as in Eq.~(\ref{eq:model}).  We now define a score function 
$\dg{\yh{n}}{\y}{\theta}$ which gives a score for assigning the labels of the segments from $\yh{n}$ to $\y$, 
{\small
\begin{eqnarray}
\dg{\yh{n}}{\y}{\theta_n} = \sum_{k \in K} \sum_{i \in \mathcal{V}} \theta_{n}^k \mathbf y^{h_nk}_i  \y_i^k 
\end{eqnarray} }
\vskip -.1in
where $K=K_s \cup K_a$. Here, $ \mathbf \theta_{n}^k$ can be interpreted as the confidence of labeling the segments of label 
$k$ correctly in the $n^{th}$ segmentation hypothesis. We want to find the labeling that maximizes the assignment score across all the segmentations. Therefore we can write inference in terms of a joint objective function as follows
{\small
\begin{eqnarray}
\label{eq:combinedObj}
\hat{\y} = \argmax_\y \max_{\yh{n} \forall h_n \in \mathcal{H}} \sum_{h_n \in \mathcal{H}} [ \df{\xh{n}}{\yh{n}}{\wh{n}} +  \dg{\yh{n}}{\y}{\theta} ]
\end{eqnarray}
}
\vskip -.1in
This formulation is equivalent to considering the labelings $y^{h_n}$ over the segmentations as unobserved variables. It is possible to use 
the latent structural SVM \cite{Yu:2009} to solve this, but it becomes intractable if the size of the segmentation hypothesis space is large. Therefore we propose an approximate two-step learning procedure to address this.
For a given set of segmentations $\mathcal{H}$, we first learn the parameters $\wh{n}$ independently 
as described in Section \ref{sec:model}.  We then train the parameters $\mathbf{\theta}$  on a separate held-out training dataset. This can now be formulated as a QP:
{\small
\begin{eqnarray}
\label{eq:model2}
\min_{\theta} & &  \frac{1}{2} \theta^T\theta - \sum_{h_n \in \mathbf{H}} \dg{\yh{n}}{\y}{\theta_n} 
s.t.  \forall k \in K : \sum_{n =1}^{|\mathbf{H}|} \theta_n^k = 1 \nonumber
\end{eqnarray}  
}
\vskip -.1in
Using the fact that  the objective function defined in Eq.~(\ref{eq:combinedObj}) is convex,
we design an iterative two-step procedure where we solve for $\yh{n}, \forall h_n \in \mathbf{H}$  in parallel 
and then solve for $\y$. This method is guaranteed to converge,
and when the number of variables scales linearly with the number of segmentation hypothesis considered, the original
problem in Eq.~(\ref{eq:combinedObj})  will become considerably slow, but our method will still scale. More formally,
we iterate between the following two problems:

{\small
\begin{minipage}{0.56\textwidth}
\begin{eqnarray}
\yhhat{n} &=& \argmax_\yh{n} \df{\xh{n}}{\yh{n}}{\wh{n}} + \dg{\yh{n}}{\hat{\y}}{\theta_n} \qquad\label{eq:inference1}
\end{eqnarray}
\end{minipage}
\begin{minipage}{0.03\textwidth}
,
\end{minipage}
\begin{minipage}{0.4\textwidth}
\begin{eqnarray}
\hat{\y} &=& \argmax_\y \dg{\yhhat{n}}{\y}{\theta_n} \label{eq:inference2}
\end{eqnarray}
\end{minipage}
}

\noindent
\textbf{High-level Activity Classification.}  For classifying the high-level activity, 
we compute the histograms of sub-activity and affordance labels and use 
them as features. Since the occlusion of objects also plays a major role in some activities, 
we capture this
by including in our feature vector, the fraction of objects that are occluded fully or partially in the temporal segments.
We then train a multi-class SVM classifier on training data using these features.

\vspace*{\sectionReduceTop}
\section{Experiments}\label{sec:experiments}
\vspace*{\sectionReduceBot}

{\bf Data.} 
We test our model on two 3D activity datasets: Cornell 60 \cite{SungICRA2012}
and Cornell 120 Activity Datasets.
The Cornell 60 Activity Dataset \cite{SungICRA2012} contains 60 3D videos of four different subjects performing 12 high-level activity classes. However, some of these activity classes contain only one sub-activity (e.g., \emph{working on a computer}, \emph{cooking (stirring)}, etc.) and do not contain object interactions (e.g., \emph{talking on couch}, \emph{relaxing on couch}). 

We introduce the Cornell 120 Activity Dataset dataset 
, which contains 
activity sequences of ten different high-level activities performed 
by four different subjects, where each high-level activity was performed three times. 
We thus have 61,585 total 3D video frames in our dataset. 
The high-level activities are: \{\emph{making cereal}, \emph{taking medicine}, \emph{stacking objects}, \emph{unstacking objects}, \emph{microwaving food}, \emph{picking objects}, \emph{cleaning objects}, \emph{taking food}, \emph{arranging objects}, \emph{having a meal}\}. The subjects were only given a high-level 
description of the task,\footnote{For example, the instructions for \emph{making cereal} were: 
1) Place bowl on table, 2) Pour cereal, 3) Pour milk. For \emph{microwaving food}, they were: 1) Open microwave door, 2) Place food inside, 3) Close microwave door.}  
and were asked to perform the activities multiple times with \emph{different} objects.
For example, the stacking and unstacking activities were performed with 
pizza boxes, plates and bowls.  
They performed the activities through a long sequence of sub-activities, which varied from subject to subject significantly in terms of length of the sub-activities, order of the sub-activities as well as in the way they executed
the task. The camera was mounted so that 
the subject was in view (although the subject may not be facing the camera), but often there were significant occlusions of the body parts.
See Fig.~\ref{fig:reachingData} 
for some examples.

We labeled our dataset with the sub-activity and the object affordance labels. 
We refer to the set of contiguous frames spanning a sub-activity as a temporal segment. Specifically, our sub-activity labels are: \{\emph{reaching}, \emph{moving}, \emph{pouring}, \emph{eating}, \emph{drinking}, \emph{opening}, \emph{placing}, \emph{closing}, \emph{scrubbing}, \emph{null}\}  and our affordance labels are: \{\emph{reachable}, \emph{movable}, \emph{pourable}, \emph{pourto}, \emph{containable}, \emph{drinkable}, \emph{openable}, \emph{placeable}, \emph{closable}, \emph{scrubbable}, \emph{scrubber}, \emph{stationary}\}. Table 1 in supplementary 
material shows the details on which sub-activities are present in each high-level activity.

\begin{table}[tb!]
\vskip -.1in
\caption{Results on Cornell 60 Activity Dataset \cite{SungICRA2012}, tested on \emph{``New Person"} data for 12 activity classes.}
\vspace*{\captionReduceBot}
 \label{tbl:cad_result}
\centering
{\scriptsize
\begin{tabular}{l|cc|cc|cc|cc|cc|cc}
\whline{1.1pt}
& \multicolumn{2}{c|}{bathroom} & \multicolumn{2}{c|}{bedroom} & \multicolumn{2}{c|}{kitchen} & \multicolumn{2}{c|}{living room} & \multicolumn{2}{c|}{office} & \multicolumn{2}{c}{Average}\\
 &   prec & rec & prec & rec  &   prec & rec  &   prec & rec   &   prec & rec  &   prec & rec   \\
\hline
Sung et.~al.~\cite{SungICRA2012} & 72.7 & \textbf{65.0} & \textbf{76.1} & 59.2 & 64.4 & 47.9 & 52.6 & 45.7 &73.8  & 59.8 & 67.9 & 55.5 \\
Our method & \textbf{88.9} &61.1 & 73.0 & \textbf{66.7} & \textbf{96.4} & \textbf{85.4} & \textbf{69.2} & \textbf{68.7} &  \textbf{76.7} & \textbf{75.0}  & \textbf{80.8} & \textbf{71.4} \\
\whline{1.1pt}
\end{tabular}
}
\vskip -.25in
\end{table}

{\bf Preprocessing.}
Given the raw data containing the color and depth values for every pixel in the video, we first tracked the human skeleton using Openni's skeleton tracker \cite{openni-tracker}
 for obtaining the locations of the various joints of the human skeleton. 
 However these values are not very accurate, as the Openni's skeleton tracker is only 
 designed to track human skeletons in clutter-free  environments and without any 
 occlusion of the body parts. 
In real-world human activity videos, some body parts are often occluded 
 and the interaction with the objects hinders accurate skeleton tracking. We show that 
  even with such noisy data, our method gets high accuracies by modeling
  the mutual context between the affordances and sub-activities.
 
We then use SIFT feature matching \cite{PeleECCV2008}, 
while enforcing depth consistency across the time frames for obtaining reliable object tracks 
in the 3D video.
In our current implementation, we need to provide the bounding 
box of the objects involved in the activity in the first frame of the video. Note that these are 
just bounding boxes and not object labels. In future, such bounding boxes of objects of interest 
can also be obtained in a preprocessing step by either running a set of object 
detectors \cite{li2010object} or some other methods that use 3D information 
\cite{martinez2010moped,rusu2007towards}.
The outputs from the skeleton and object tracking along with the RGBD videos 
are then used to generate the features.


\textbf{Labeling results on the Cornell 60 Activity Dataset.} Table \ref{tbl:cad_result} shows the precision and recall of the high-level activities on the Cornell Activity Dataset \cite{SungICRA2012}. Following Sung et.~al.'s \cite{SungICRA2012} experiments, we considered the same five groups of activities based on their location, and learnt a separate model 
for each location. To make it a fair comparison, we do not assume perfect segmentation of sub-activities and do not use any object information. Therefore, we train our model with only sub-activity nodes and consider segments of uniform size (20 frames per segments). We consider only a subset of our features described in Section \ref{sec:model} that are possible to compute from the tracked human skeleton and RGBD data provided in this dataset. 
Table \ref{tbl:cad_result} shows that our model significantly outperforms Sung et.~al.'s MEMM model even when using only the sub-activity nodes and a simple segmentation algorithm. 
More detailed results are provided in the supplementary material.


\begin{table*}[tb!]
\caption{{\bf Results on Cornell 120 Activity Dataset}, showing average micro precision/recall, and average macro precision and recall for affordance, sub-activities and high-level activities.
Standard error is also reported.}
\vspace*{\captionReduceBot}
 \label{tbl:labeling_result}
{\footnotesize
\newcolumntype{P}[2]{>{\footnotesize#1\hspace{0pt}\arraybackslash}p{#2}}
\setlength{\tabcolsep}{2pt}
\centering
\resizebox{\hsize}{!}
 {
\begin{tabular}
{@{}p{0.25\linewidth} |P{\centering}{15mm}P{\centering}{15mm}P{\centering}{15mm}|P{\centering}{15mm}P{\centering}{15mm}P{\centering}{15mm}| P{\centering}{15mm}P{\centering}{15mm}P{\centering}{15mm}@{}}
\whline{1.1pt}
 & \multicolumn{3}{c|}{Object Affordance} & \multicolumn{3}{c|}{Sub-activity} & \multicolumn{3}{c}{High-level Activity} \\
\cline{2-10}
 & \multicolumn{1}{c}{micro} & \multicolumn{2}{c|}{macro} & \multicolumn{1}{c}{micro} &  \multicolumn{2}{c|}{macro}  & \multicolumn{1}{c}{micro} &  \multicolumn{2}{c}{macro}  \\
\whline{0.4pt}
     method  & $P/R$ & Prec.  & \multicolumn{1}{c|}{Recall} &  $P/R$ & Prec. &  \multicolumn{1}{c|}{Recall} & P/R & Prec. &  \multicolumn{1}{c}{Recall}  \\
\whline{0.8pt}
 \emph{max class}   & 65.7 $\pm$ 1.0 & 65.7 $\pm$ 1.0 & 8.3 $\pm$ 0.0 & 29.2 $\pm$ 0.2  & 29.2 $\pm$ 0.2  & 10.0 $\pm$ 0.0 & 10.0 $\pm$ 0.0 & 10.0 $\pm$ 0.0   & 10.0 $\pm$ 0.0   \\
\emph{image only}                   & 74.2 $\pm$ 0.7 & 15.9 $\pm$ 2.7 & 16.0 $\pm$ 2.5 & 56.2 $\pm$ 0.4 & 39.6 $\pm$ 0.5 & 41.0 $\pm$ 0.6 & 34.7 $\pm$ 2.9 & 24.2 $\pm$ 1.5 & 35.8 $\pm$ 2.2  \\
\emph{SVM multiclass}                      & 75.6 $\pm$ 1.8 & 40.6 $\pm$ 2.4 & 37.9 $\pm$ 2.0 & 58.0 $\pm$ 1.2 & 47.0 $\pm$ 0.6 & 41.6 $\pm$ 2.6 & 30.6 $\pm$ 3.5 & 27.4 $\pm$ 3.6 & 31.2 $\pm$ 3.7 \\
\emph{MEMM}\cite{SungICRA2012}                      &  - & - & - &  -  & - & - & 26.4 $\pm$ 2.0 & 23.7 $\pm$ 1.0 & 23.7 $\pm$ 1.0 \\
\whline{0.6pt}
\emph{object only}                           & 86.9 $\pm$ 1.0 & 72.7 $\pm$ 3.8 & 63.1 $\pm$ 4.3 & - & - & -                                        & 59.7 $\pm$ 1.8 & 56.3 $\pm$ 2.2 & 58.3 $\pm$ 1.9  \\
\emph{sub-activity only}                     & - & -  & - &                                      71.9 $\pm$ 0.8  & 60.9 $\pm$ 2.2 & 51.9 $\pm$ 0.9 & 27.4 $\pm$ 5.2 & 31.8 $\pm$ 6.3 & 27.7 $\pm$ 5.3  \\
\emph{no temporal interactions}              & 87.0 $\pm$ 0.8 & 79.8 $\pm$ 3.6 & 66.1 $\pm$ 1.5 & 76.0 $\pm$ 0.6 & 74.5 $\pm$ 3.5 & 66.7 $\pm$ 1.4 & 81.4 $\pm$ 1.3 & 83.2 $\pm$ 1.2 & 80.8 $\pm$ 1.4 \\ 
\emph{no object interactions}                & 88.4 $\pm$ 0.9 & 75.5 $\pm$ 3.7 & 63.3 $\pm$ 3.4 & 85.3 $\pm$ 1.0 & 79.6 $\pm$ 2.4 & 74.6 $\pm$ 2.8 & 80.6 $\pm$ 2.6 & 81.9 $\pm$ 2.2 & 80.0 $\pm$ 2.6 \\
\emph{full model: groundtruth seg}                           & \textbf{91.8} $\pm$ 0.4 & \textbf{90.4} $\pm$ 2.5 & \textbf{74.2} $\pm$ 3.1 & \textbf{86.0} $\pm$ 0.9 & \textbf{84.2} $\pm$ 1.3 & \textbf{76.9} $\pm$ 2.6 & \textbf{84.7} $\pm$ 2.4 & \textbf{85.3} $\pm$ 2.0 & \textbf{84.2} $\pm$ 2.5   \\
\hline
\multicolumn{10}{c}{}\\
\multicolumn{10}{c}{Full model. End-to-end results, \emph{without} assuming any ground-truth temporal segmentation is given.}\\
\hline
\emph{full, 1 segment.~(best)} 			& 83.1 $\pm$ 1.1 & 70.1 $\pm$ 2.3 & 63.9 $\pm$ 4.4 & 66.6 $\pm$ 0.7 & 62.0 $\pm$ 2.2 & 60.8 $\pm$ 4.5 & 77.5 $\pm$ 4.1 & 80.1 $\pm$ 3.9 & 76.7 $\pm$ 4.2  \\
\emph{full, 1 segment.~(averaged)} 		& 81.3 $\pm$ 0.4 & 67.8 $\pm$ 1.1 & 60.0 $\pm$ 0.8 & 64.3 $\pm$ 0.7 & 63.8 $\pm$ 1.1 & 59.1 $\pm$ 0.5 & 79.0 $\pm$ 0.9 & 81.1 $\pm$ 0.8 & 78.3 $\pm$ 0.9  \\
\emph{full, multi-seg learning }         & \textbf{83.9} $\pm$ 1.5 & \textbf{75.9} $\pm$ 4.6 & \textbf{64.2} $\pm$ 4.0 & \textbf{68.2} $\pm$ 0.3 & \textbf{71.1} $\pm$ 1.9 & \textbf{62.2} $\pm$ 4.1 & \textbf{80.6} $\pm$ 1.1  & \textbf{81.8} $\pm$ 2.2 & \textbf{80.0} $\pm$  1.2 \\
\whline{1.1pt}
\end{tabular}
}
}
 \vskip -.1in
\end{table*}

\begin{figure}[tb!]
 \centering 
 \begin{minipage}[t]{0.34\linewidth}
\centering
  \includegraphics[width=\linewidth,height=1.05in]{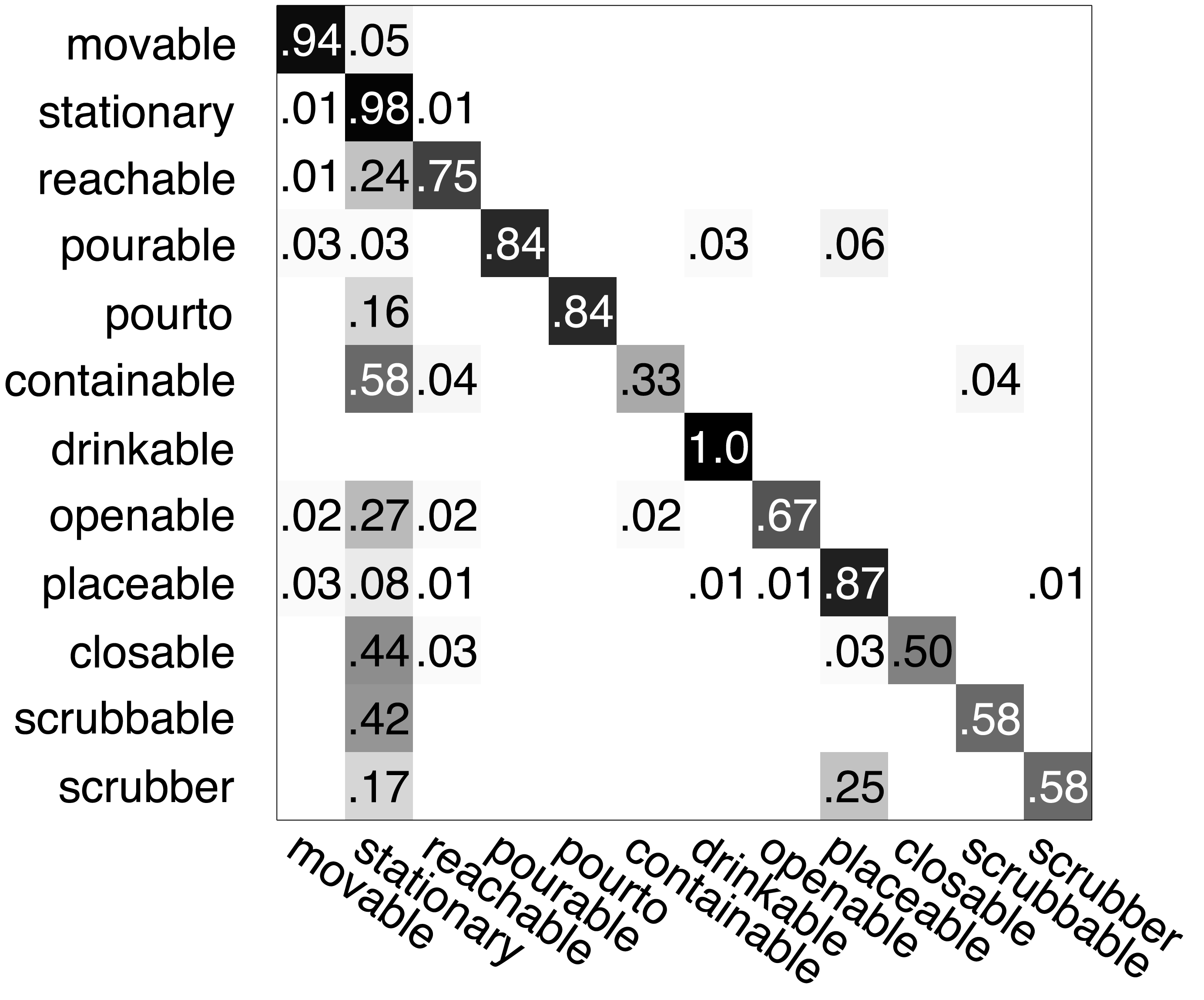}
\end{minipage}
 \begin{minipage}[t]{0.3\linewidth}
 \vskip -1.05in
\centering 
 \includegraphics[width=\linewidth,height=1in]{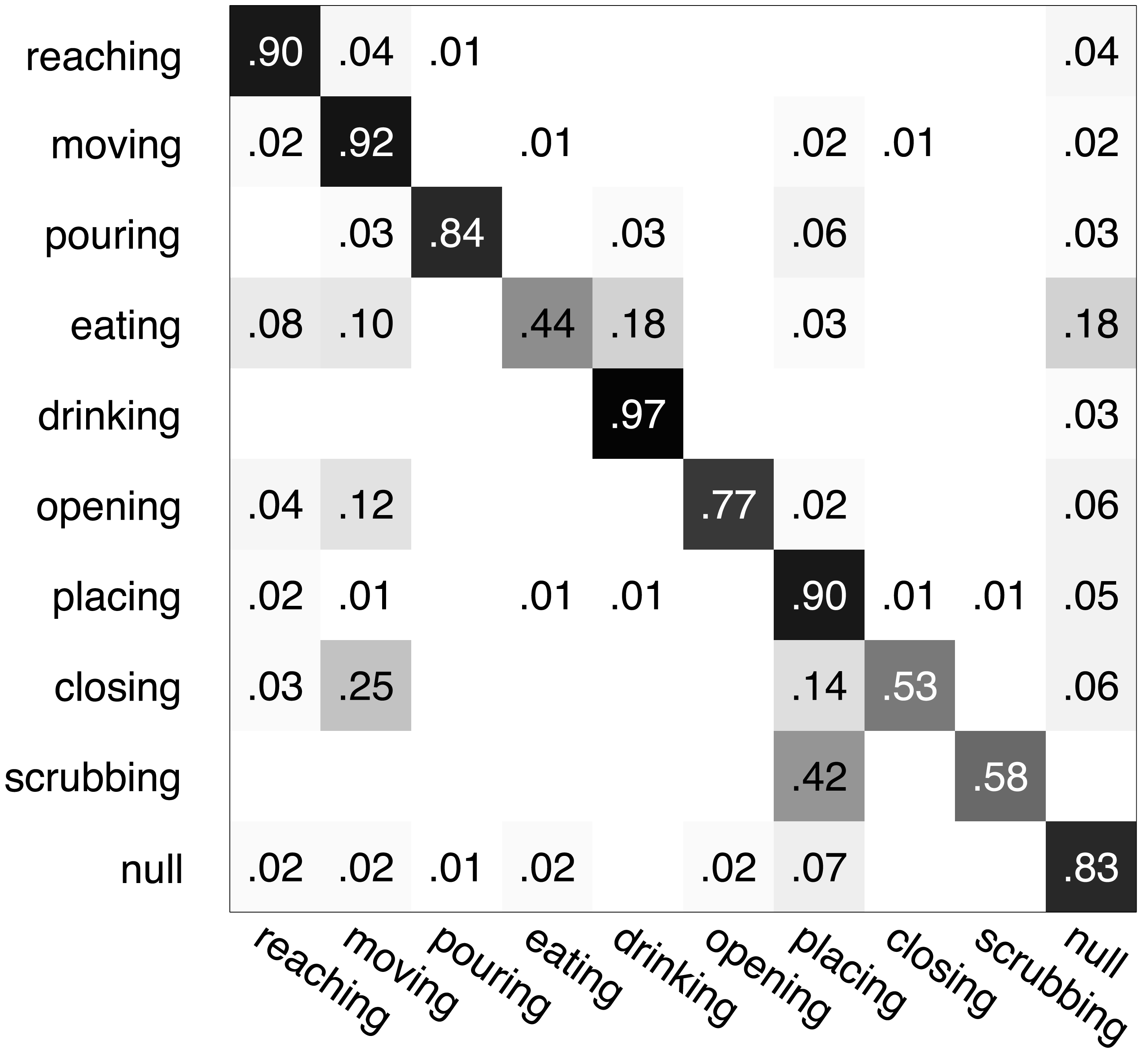}
 \end{minipage}
 \begin{minipage}[t]{0.34\linewidth}
\vskip -1.05in
\centering
 \includegraphics[width=\linewidth,height=1.08in]{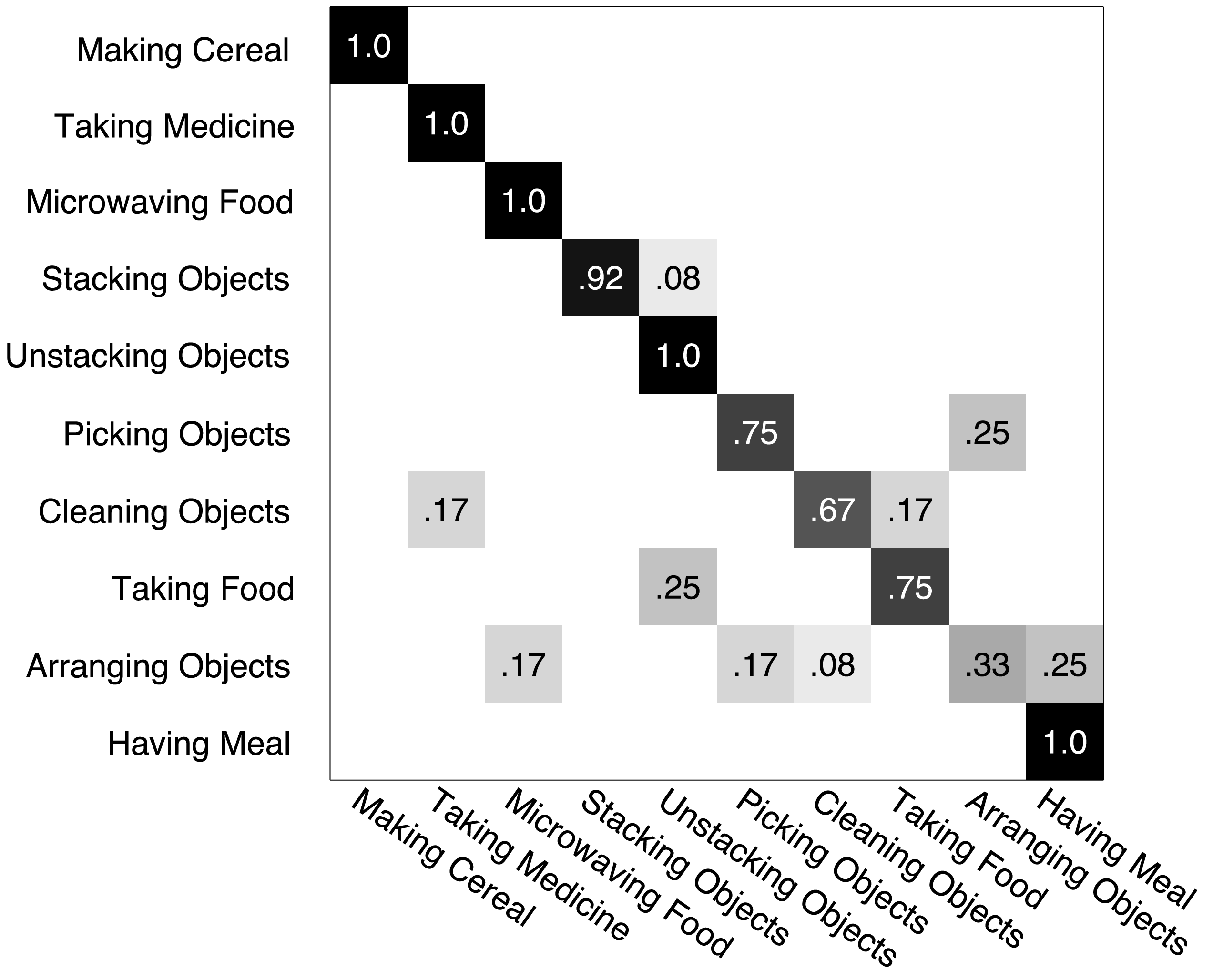}
 \end{minipage}
  \vspace*{\captionReduceTop}
 \caption{Confusion matrix for affordance labeling (left), sub-activity labeling (middle) and high-level
  activity labeling (right) of the test RGB-D videos.}
 \vspace*{\captionReduceBot}
 \label{fig:ActivityconfusionMatix}
 \vskip -.05in
 \end{figure}

{\bf Labeling results on Cornell 120 Activity Dataset.}
Table \ref{tbl:labeling_result} shows the performance of various models on object affordance, sub-activity and high-level activity labeling. 
These results are obtained using 4-fold cross-validation and averaging performance across 
the folds. Each fold constitutes the activities performed by one subject, therefore the model is trained on activities of 
three subjects and \emph{tested on a new subject}. We report both the micro and macro averaged precision and recall over various classes along with standard error. Since our algorithm can only predict one label for each segment, micro precision and recall are same as the percentage of correctly classified segments. Macro precision and recall are  the averages of precision and recall respectively for all classes.

\begin{figure*}[t!]
\vskip -.1in
\begin{minipage}[t]{0.16\linewidth}
\centering
\includegraphics[width=\linewidth,height=0.6in]{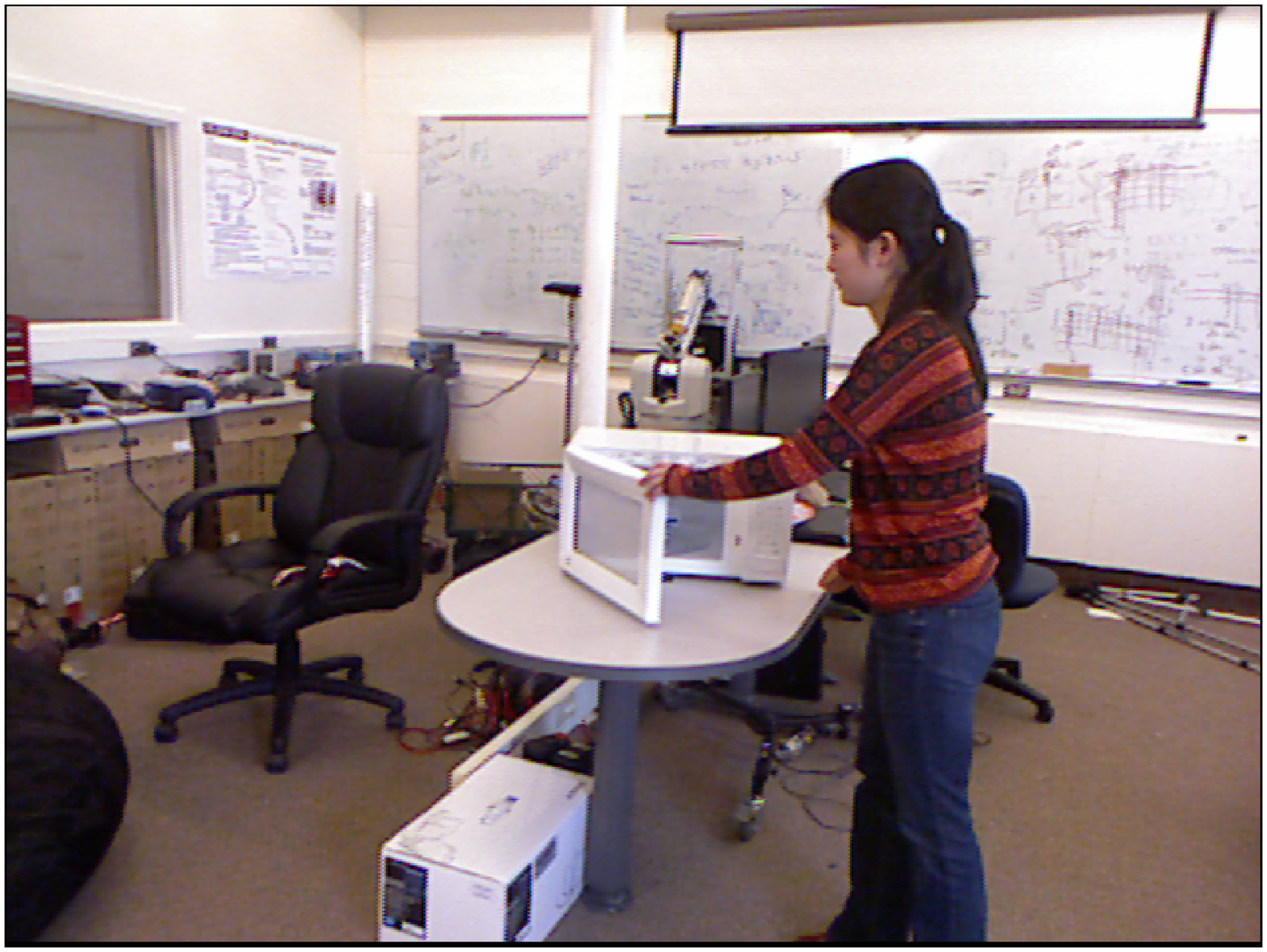}
{\scriptsize Subject \emph{opening}  \emph{openable} object1}
\end{minipage}
\begin{minipage}[t]{0.16\linewidth}
\centering
\includegraphics[width=\linewidth,height=0.6in]{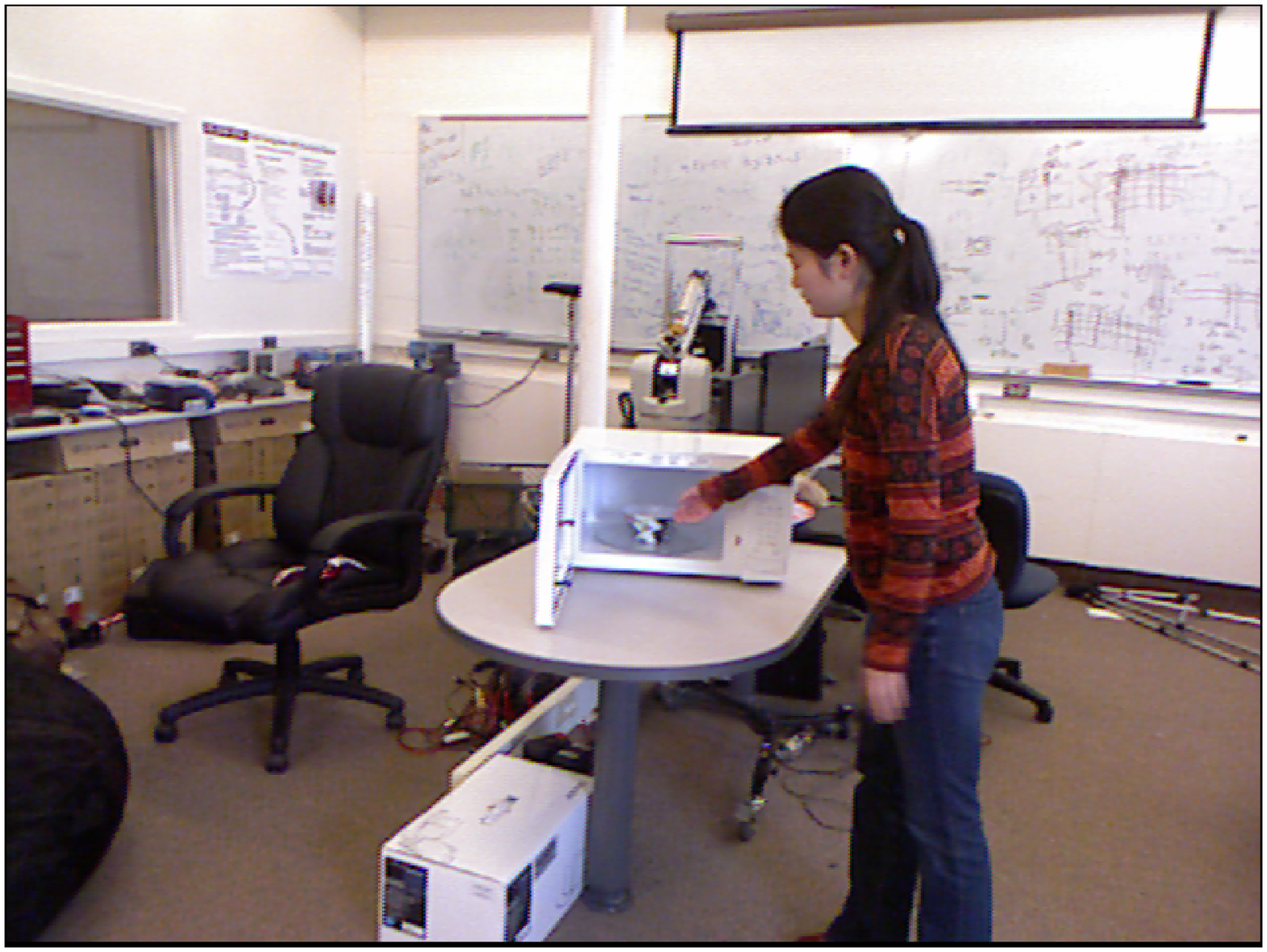}
{\scriptsize Subject \emph{reaching}  \emph{reachable} object2}

\end{minipage}
\begin{minipage}[t]{0.16\linewidth}
\centering
\includegraphics[width=\linewidth,height=0.6in]{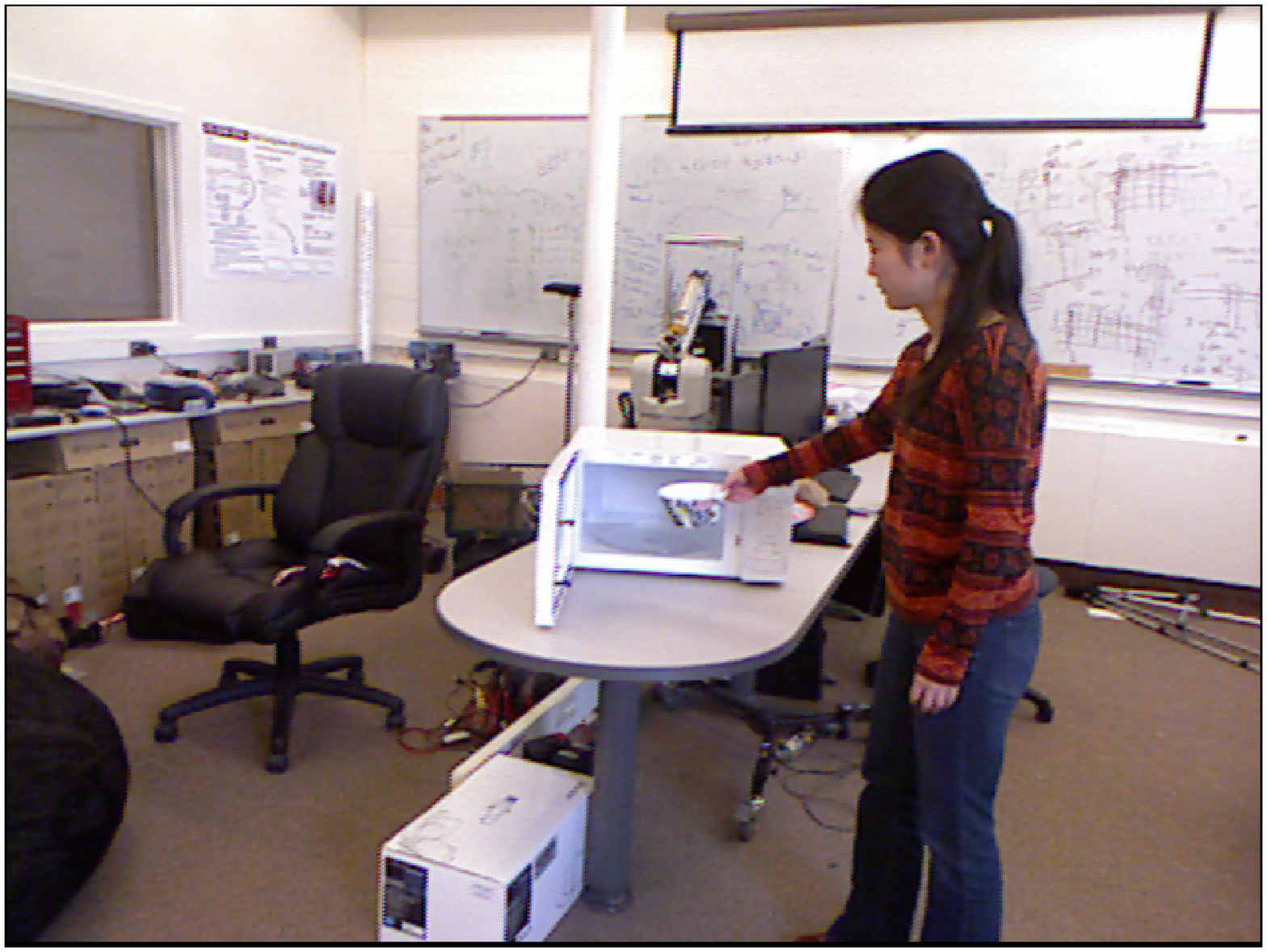}
{\scriptsize Subject \emph{moving} \emph{movable} object2}

\end{minipage}
\begin{minipage}[t]{0.16\linewidth}
\centering
\includegraphics[width=\linewidth,height=0.6in]{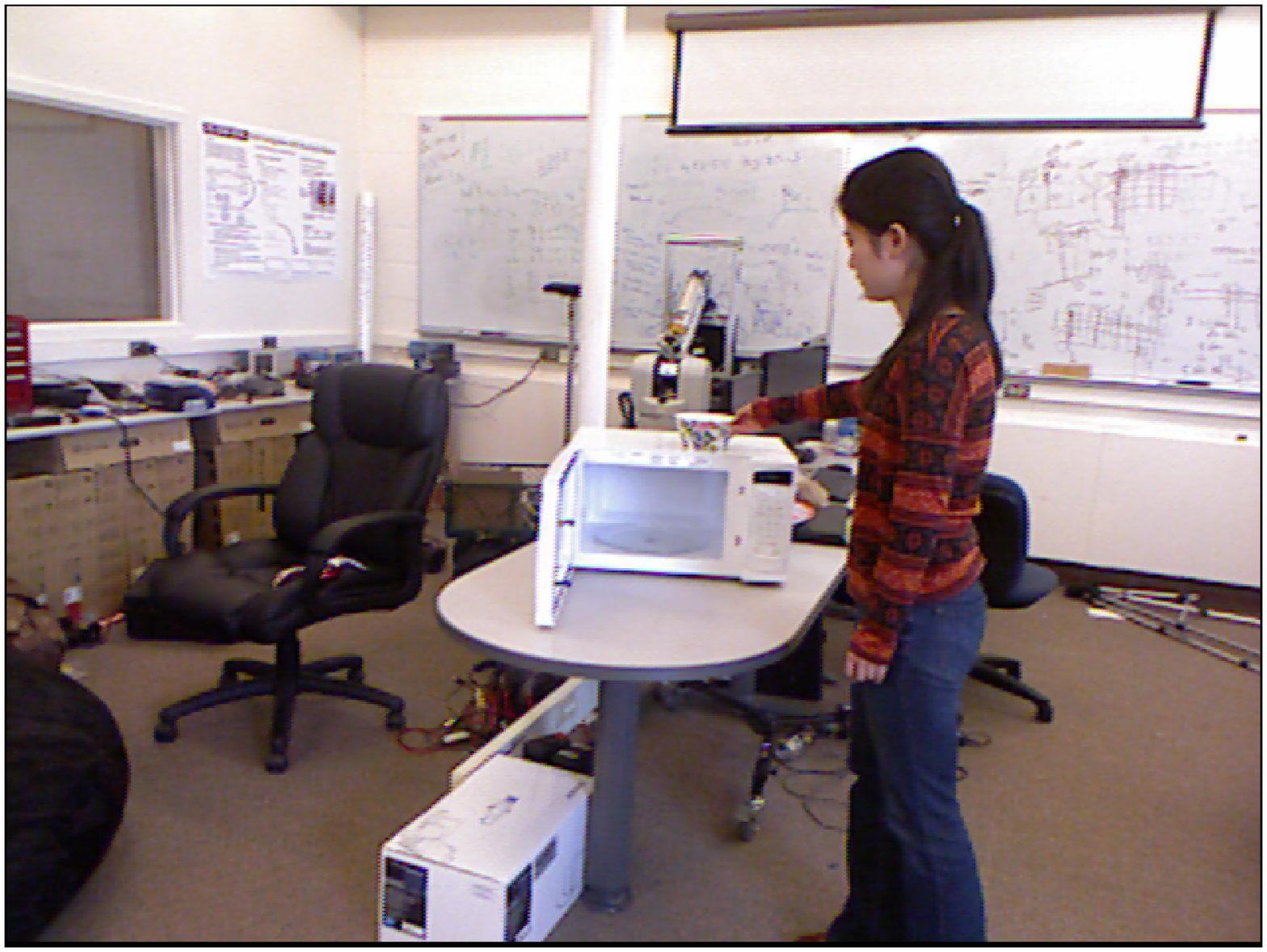}
{\scriptsize Subject \emph{placing}  \emph{placable} object2}
\end{minipage}
\begin{minipage}[t]{0.16\linewidth}
\centering
\includegraphics[width=\linewidth,height=0.6in]{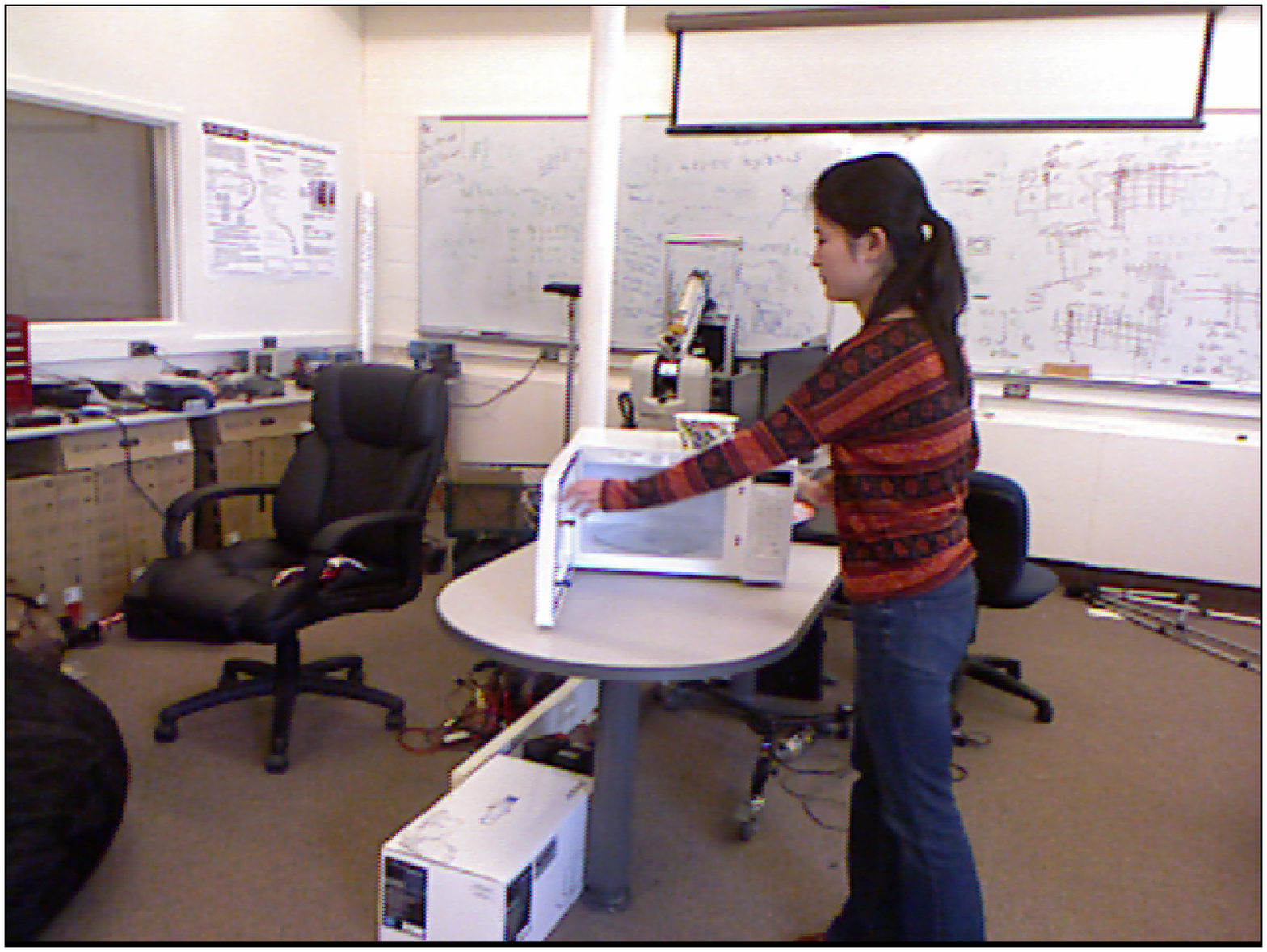}
{\scriptsize Subject \emph{reaching}  \emph{reachable} object1}
\end{minipage}
\begin{minipage}[t]{0.15\linewidth}
\centering
\includegraphics[width=\linewidth,height=0.6in]{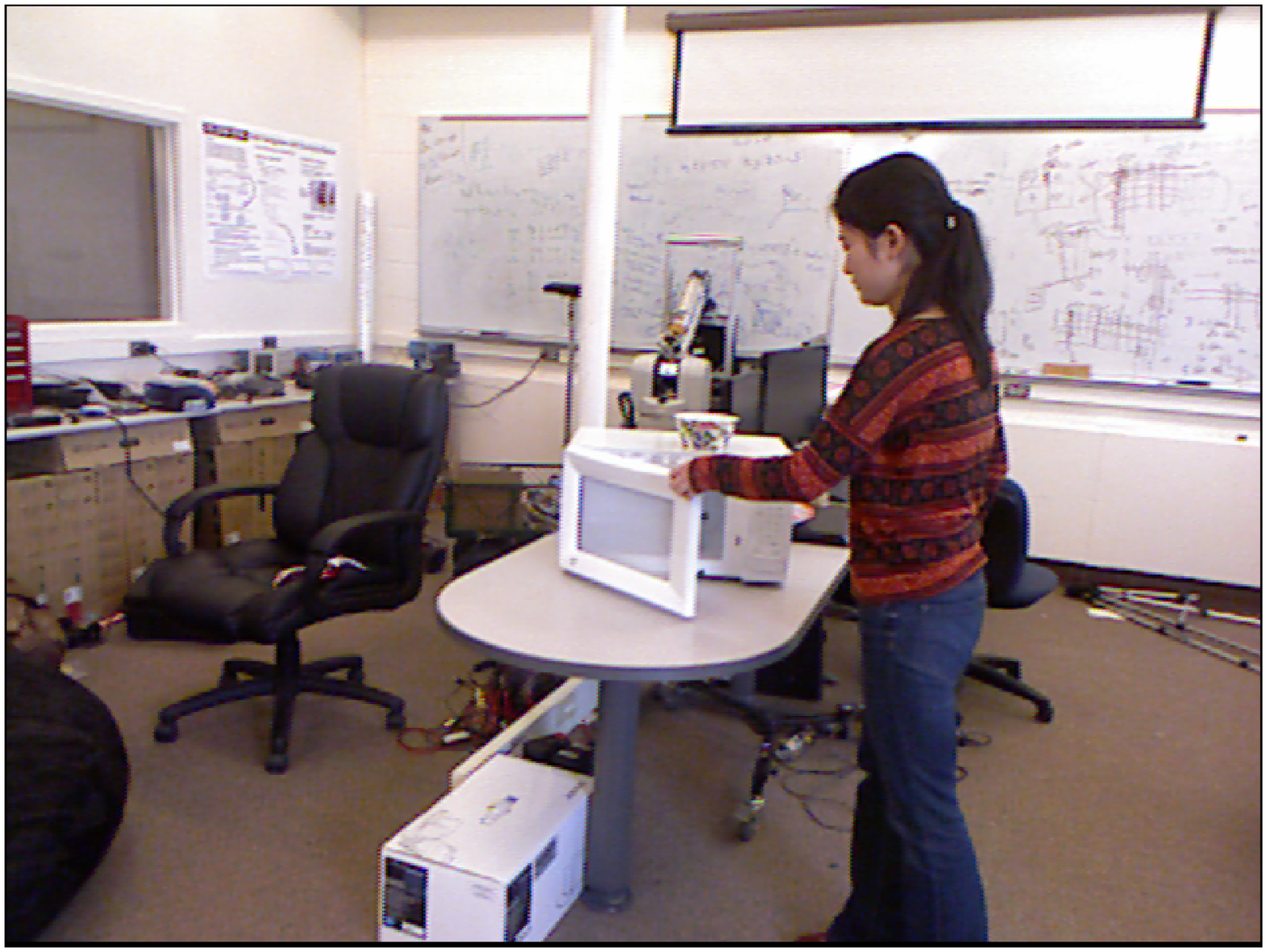}
{\scriptsize Subject \emph{closing}  \emph{closable} object1}
\end{minipage}
\vspace*{\captionReduceTop}
 \caption{Output of our algorithm: Sequence of images from the \emph{taking food} activity labeled with sub-activity and object affordance labels.} 
\vspace*{\captionReduceBot}
\vspace*{\captionReduceBot}
\label{fig:labelingresult}
 \end{figure*}

Assuming ground-truth temporal segmentation is given, the results for our \emph{full model} are
shown in Table~\ref{tbl:labeling_result} on line 9, its variations on lines 5-8 and the baselines on lines 1-3.
 The results in lines 10-12 correspond to the case when temporal segmentation is not assumed. 
In comparison to a basic SVM multiclass model \cite{joachims2009cutting} ( referred to as \emph{SVM multiclass} when using all features and \emph{image only} when using only image features), which is equivalent to only considering the nodes in our MRF without any edges, our model performs significantly better.
 We also compare with the high-level activity classification results obtained from the method presented in 
 \cite{SungICRA2012}. We ran their code on our dataset and obtain accuracy of 26.4\%, whereas our method gives  an 
 accuracy of 84.7\% when ground truth segmentation is available and 80.6\% otherwise.   
Figure \ref{fig:labelingresult} shows a sequence of images from \emph{taking food} activity along with the inferred labels.
Figure \ref{fig:ActivityconfusionMatix} shows the confusion matrix for labeling affordances, sub-activities and high-level 
activities with our proposed method. We can see that there is a strong diagonal with a few errors such as \emph{scrubbing} misclassified as \emph{placing}, and \emph{picking objects} misclassified as \emph{arranging objects}.

 We analyze our model to gain insight into which interactions provide useful information 
 by comparing our full model to variants of our model.

{\bf How important is object context for activity detection?} We show the importance of object context for sub-activity labeling by learning a variant of our model without the object nodes (referred to as \emph{sub-activity only}).   With object context, the micro precision increased by 14.1\% and both macro precision and recall increased by around 23.3\% over \emph{sub-activity only}.
Considering object information (affordance labels and occlusions) also improved the high-level 
activity accuracy by 3-fold.    

{ \bf How important is activity context for affordance detection?} We also show the importance of context from sub-activity for affordance detection by learning our model without the sub-activity nodes (referred to as \emph{object only}).  With sub-activity context, the micro precision increased by 4.9\% and the macro precision and recall increased by 17.7\%
and 11.1\% respectively for affordance labeling over \emph{object only}.
The relative gain is less compared to that obtained in sub-activity detection as the \emph{object only} 
model still has object-object context which helps in affordance detection.    

{\bf How important is object - object context for affordance detection?}  In order to study the 
effect of the object-object  interactions for affordance detection, we learnt our model without the object-object edge potentials (referred to as \emph{no object interactions}). We see a considerable improvement in affordance detection when the object interactions are modeled, the macro recall increased by 14.9\% and the macro precision by about 10.9\%. This shows that sometimes just the context from the human activity alone is not sufficient to determine the affordance of an object.   

{ \bf How important is temporal context?} We also learn our model without the temporal edges (referred to as \emph{no temporal interactions}). Modeling temporal interactions increased the micro precision by 4.8\% and 10.0\% for affordances and sub-activities respectively and increased the micro precision for high-level activity by 3.3\%.

{\bf End-to-end Results.}
Given the RGB-D video, we obtain the final labeling using our method described in 
Section \ref{sec:segmentation}. To generate the segmentation hypothesis set $\mathcal{H}$ we consider three different segmentation algorithms, and generate multiple segmentations by changing 
their parameters.
 We consider: 1) uniform segmentation parameterized by segment size, 2) graph based segmentation \cite{Felzenszwalb:2004} with edge weights defined using the displacement of skeletal joints, and 3) graph based segmentation with edge weights defined using rate of change of skeletal joint locations. The lines 10-12 of Table \ref{tbl:labeling_result} show the results of the best performing segmentation,
average performance all the segmentations considered, and our proposed method for combining the segmentations. 
We see that our method improves the performance over considering a single best performing segmentation: macro precision increased by 5.8\% and 9.1\% for affordance and sub-activity labeling respectively.

\vspace*{\sectionReduceTop}
\section{Conclusion}
\vspace*{\sectionReduceBot}

In this paper, we considered the task of jointly labeling human activities and 
object affordances from RGB-D videos. The activities we consider happen
over a long time period, and comprise several sub-activities performed in a sequence. 
We formulated
this problem as a Markov Random Field, and learned the parameters of the
model using a structural SVM formulation. Our model also incorporates the
temporal segmentation problem by computing several segmentations and
considering 
labeling 
over these segmentations as
latent variables. In extensive experiments over a challenging dataset, 
we show that our method 
achieves an end-to-end accuracy precision of 81.8\% and
recall of 80.0\% for labeling the activities performed by a different
 subject than the ones in the training set. We also showed that
 it is important to model the different properties (object affordances,
 object-object interaction, temporal interactions, etc.)  in order to 
 achieve good performance.



\newpage
\setlength{\bibsep}{1pt}
{ 
  \small
   \bibliographystyle{unsrt}
\bibliography{references}
}

\end{document}